\pdfoutput=1

\documentclass[11pt]{article}
\usepackage[dvipsnames, svgnames, x11names]{xcolor}
\usepackage{ACL2023}

\usepackage{times}
\usepackage{latexsym}

\usepackage[T1]{fontenc}

\usepackage[utf8]{inputenc}

\usepackage{microtype}
\usepackage{inconsolata}

\usepackage{booktabs}
\usepackage{multirow}
\usepackage{amsmath}
\usepackage{amssymb}
\usepackage{xcolor}
\usepackage{pifont}

\usepackage{tikz}
\usepackage{subfigure}
\usepackage{pgfplots}
\usepackage{makecell,rotating,diagbox}
\usepackage{verbatim}
\usepackage{colortbl}

\usepackage{algorithm}
\usepackage{algorithmic}
\usepackage{graphicx}
\usepackage{fancybox}
\usepackage[normalem]{ulem}
\usepackage{wasysym}

\usetikzlibrary{plotmarks}
\usetikzlibrary{patterns}
\usetikzlibrary{pgfplots.groupplots}

\usetikzlibrary{plotmarks}
\usetikzlibrary{arrows}
\usetikzlibrary{decorations}
\usetikzlibrary{decorations.pathreplacing}
\usetikzlibrary{decorations.markings}
\usetikzlibrary{backgrounds}
\usetikzlibrary{fit}
\usetikzlibrary{positioning}
\usetikzlibrary{calc}
\usetikzlibrary{patterns}
\usetikzlibrary{shadows}
\usetikzlibrary[shapes.multipart]
\usetikzlibrary{pgfplots.groupplots}

\newdimen\legendmargin 
\newdimen\legendwidth 
\newdimen\legendsep 

\definecolor{ugreen}{cmyk}{1,0,1,0.498}
\definecolor{lyyblue}{cmyk}{0.8278,0.3333,0,0.2941}
\definecolor{lyygreen}{cmyk}{0.6813,0,0.725,0.3725}
\definecolor{lyyred}{cmyk}{0,0.8855,0.8767,0.1098}
\definecolor{dblue}{cmyk}{1,0.5487,0,0.5569}
\definecolor{lypurple}{HTML}{e0c2c0}
\definecolor{lygreen}{HTML}{eff67b}
\definecolor{lyblue}{HTML}{d5ddef}
\definecolor{lyyellow}{HTML}{fdfab5}
\definecolor{lypink}{HTML}{ffe0db}
\definecolor{lyred}{HTML}{b71a3b}
\definecolor{lygrey}{HTML}{c4c0c2}
\definecolor{lyorange}{HTML}{eab586}

\newcommand{\tab}[1]{Table \ref{#1}}%
\newcommand{\fig}[1]{Fig. \ref{#1}}%
\newcommand{\eqn}[1]{Eq. \ref{#1}}%

\usepackage{mathtools}
\DeclarePairedDelimiter{\norm}{\lVert}{\rVert} 

\title{Understanding Parameter Sharing in Transformers}

\author{
  Ye Lin\textsuperscript{1},
  Mingxuan Wang\textsuperscript{2},
  Zhexi Zhang\textsuperscript{2},
  Xiaohui Wang\textsuperscript{2},
  Tong Xiao\textsuperscript{1,3},
  Jingbo Zhu\textsuperscript{1,3} \\
  \textsuperscript{1}NLP Lab, School of Computer Science and Engineering, \\
    Northeastern University, Shenyang, China \\
  \textsuperscript{2} ByteDance \\
  \textsuperscript{3}NiuTrans Research, Shenyang, China \\
  {\tt \{linye2015\}@outlook.com}\\
  {\tt \{wangxiaohui.neo,zhangzhexi,wangmingxuan.89\}@bytedance.com}\\
  {\tt \{xiaotong,zhujingbo\}@mail.neu.edu.cn} \\
}

\begin{document}
\maketitle
\begin{abstract}

Parameter sharing has proven to be a parameter-efficient approach. 
Previous work on Transformers has focused on sharing parameters in different layers, which can improve the performance of models with limited parameters by increasing model depth.
In this paper, we study why this approach works from two perspectives.
First, increasing model depth makes the model more complex, and we hypothesize that the reason is related to \textbf{model complexity} (referring to FLOPs).
Secondly, since each shared parameter will participate in the network computation several times in forward propagation, its corresponding gradient will have a different range of values from the original model, which will affect the model convergence.
Based on this, we hypothesize that \textbf{training convergence} may also be one of the reasons.
Through further analysis, we show that the success of this approach can be largely attributed to better convergence, with only a small part due to the increased model complexity.
Inspired by this, we tune the training hyperparameters related to model convergence in a targeted manner.
Experiments on 8 machine translation tasks show that our model achieves competitive performance with only half the model complexity of parameter sharing models.


\end{abstract}

\section{Introduction}

As a simple and effective strategy in parameter-efficient research, parameter sharing has been extensively studied in recent years \cite{DBLP:conf/iclr/UllrichMW17,DBLP:conf/cvpr/GaoDH19,DBLP:conf/naacl/HaoWYWZT19,DBLP:journals/corr/abs-2106-10002}.  
Previous work on Transformer has focused on sharing parameters along model depth for better performance with limited parameters \cite{DBLP:conf/iclr/DehghaniGVUK19,DBLP:journals/corr/abs-2108-10417,DBLP:journals/corr/abs-2104-06022,DBLP:conf/iclr/LanCGGSS20}. 

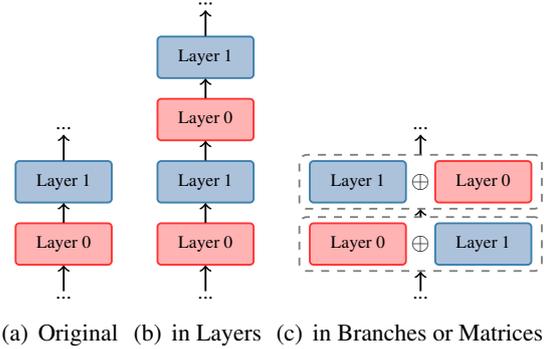
\begin{figure}
  \centering
  \subfigure[Original]{
    \tikzstyle{layer} = [rectangle,line width = 0.7pt,rounded corners=0.05cm,minimum width=2.3cm,minimum height=1cm,inner sep=0.1cm,font=\large]
    \tikzstyle{conect} = [->,line width = 0.7pt]
    
    \begin{tikzpicture}[node distance = 0,scale = 0.55]
    \tikzstyle{every node}=[scale=0.55]
    \node(layer1)[layer,draw=red!80,fill=red!30]{Layer 0};
    \node(layer2)[layer,above of = layer1,yshift=1.5cm,draw=lyyblue!80,fill=lyyblue!30]{Layer 1};
    \draw[conect](layer1.north)to(layer2.south);
    \node(point1)[below of = layer1,yshift=-1.3cm,font=\Large]{...};
    \node(point2)[above of = layer2,yshift=1.3cm,font=\Large]{...};
    \draw[conect](point1.north)to(layer1.south);
    \draw[conect](layer2.north)to(point2.south);
    \end{tikzpicture}
  \label{}
  } 
  \subfigure[in Layers]{
    \tikzstyle{layer} = [rectangle,line width = 0.7pt,rounded corners=0.05cm,minimum width=2.3cm,minimum height=1cm,inner sep=0.1cm,font=\large]
    \tikzstyle{conect} = [->,line width = 0.7pt]
    
    \begin{tikzpicture}[node distance = 0,scale = 0.55]
    \tikzstyle{every node}=[scale=0.55]
    \node(layer1)[layer,draw=red!80,fill=red!30]{Layer 0};
    \node(layer2)[layer,above of = layer1,yshift=1.5cm,draw=lyyblue!80,fill=lyyblue!30]{Layer 1};
    \node(layer3)[layer,above of = layer2,yshift=1.5cm,draw=red!80,fill=red!30]{Layer 0};
    \node(layer4)[layer,above of = layer3,yshift=1.5cm,draw=lyyblue!80,fill=lyyblue!30]{Layer 1};
    \draw[conect](layer1.north)to(layer2.south);
    \draw[conect](layer2.north)to(layer3.south);
    \draw[conect](layer3.north)to(layer4.south);
    \node(point1)[below of = layer1,yshift=-1.3cm,font=\Large]{...};
    \node(point2)[above of = layer4,yshift=1.3cm,font=\Large]{...};
    \draw[conect](point1.north)to(layer1.south);
    \draw[conect](layer4.north)to(point2.south);
    \end{tikzpicture}
  \label{}
  }
  \subfigure[in Branches or Matrices]{
    \tikzstyle{layer} = [rectangle,line width = 0.7pt,rounded corners=0.05cm,minimum width=2.3cm,minimum height=1cm,inner sep=0.1cm,font=\large]
    \tikzstyle{frame} = [rectangle,line width = 0.7pt,dashed,rounded corners=0.1cm,minimum width=5.8cm,minimum height=1.3cm, draw=gray]
    \tikzstyle{conect} = [->,line width = 0.7pt]
    
    \begin{tikzpicture}[node distance = 0,scale = 0.55]
    \tikzstyle{every node}=[scale=0.55]
    \node(layer1_1)[layer,draw=red!80,fill=red!30]{Layer 0};
    \node()[right of = layer1_1,xshift=1.5cm,scale=1.5]{$\oplus$};
    \node(frame1)[frame,above of = layer1_1,xshift=1.5cm]{};
    \node(layer1_2)[layer,right of = layer1_1,xshift=3cm,draw=lyyblue!80,fill=lyyblue!30]{Layer 1};
    \node(layer2_1)[layer,above of = layer1,yshift=1.5cm,draw=lyyblue!80,fill=lyyblue!30]{Layer 1};
    \node()[right of = layer2_1,xshift=1.5cm,scale=1.5]{$\oplus$};
    \node(frame2)[frame,above of = layer2_1,xshift=1.5cm]{};
    \node(layer2_2)[layer,right of = layer2_1,xshift=3cm,draw=red!80,fill=red!30]{Layer 0};
    \draw[conect](frame1.north)to(frame2.south);
    \node(point1)[below of = frame1,yshift=-1.3cm,font=\Large]{...};
    \node(point2)[above of = frame2,yshift=1.3cm,font=\Large]{...};
    \draw[conect](point1.north)to(frame1.south);
    \draw[conect](frame2.north)to(point2.south);
    \end{tikzpicture}
  \label{}
  }
  \caption{Comparison of different parameter sharing methods (Different colors denote parameters from different sources. $\oplus$ denotes the average/concatenation operation when sharing in branches/matrices. ).}
  \label{fig:comparison}
\end{figure}

While sharing parameters to stack more layers achieves better performance than not sharing, there hasn't been a clear explanation of why it works.
In this paper, we hypothesize that the reason may come from two perspectives: 1) model complexity and  2) training convergence.
First, to explore the impact of complex structures on performance, we propose increasing model complexity by sharing parameters in two other dimensions: model branches and weight matrices.
The comparison of these three structures can be seen in \fig{fig:comparison}.
Experiments in Section \ref{sec:complexity} show that all three models can achieve better performance than models without parameter sharing, which shows that model complexity does have a positive impact on performance with limited performance.
However, so far, we have not been able to quantify the impact of model complexity, because the above experiments will affect both model complexity and convergence.


Through further analysis in Section \ref{sec:convergence}, we verify the phenomenon that the parameter sharing model converges faster and better than the original model.
To achieve a similar convergence state, we tune training hyperparameters, i.e., the initial learning rate, warmup steps, training batches, and training steps, in a targeted manner.
In addition, $L_2$ regularization is applied for training stability. 
Our experiments show that the simpler original model with hyperparameter tuning can achieve competitive performance ($\Delta\mathrm{BLEU}<0.5$, the gap can be attributed to model complexity) with only half the complexity of parameter sharing models.

To summarize, our contributions are as follows:
\begin{itemize}
  \item In this paper, we study the parameter sharing approach from two perspectives: model complexity and training convergence. 
  We find that its success depends largely on the optimized training convergence and a small part on the increased model complexity.
  \item For better convergence, we tune the related training hyperparameters in a targeted manner.
  Our experiments on 8 WMT tasks show that the original model can achieve competitive performance against the sharing model, but with only half the model complexity.
  \item By tuning hyperparameters, our original Transformer model can achieve 1.00 higher BLEU points on average (e.g., a BLEU score of 28.62/29.87 of Transformer-base/big on WMT14 English-German).
  It inspires us that the original Transformer model still has room for improvement, and studies on efficient methods need a strong baseline.
\end{itemize}

\section{Background}
\subsection{Transformer}
\label{sub-sec:transformer}

In this work, we choose Transformer for study because it is one of the state-of-the-art neural networks in many sequence learning tasks \cite{DBLP:conf/nips/VaswaniSPUJGKP17,DBLP:conf/acl/WangLXZLWC19,DBLP:journals/corr/abs-2008-07772}. 
The Transformer model consists of multiple stacked layers, mainly including the embedding layer, the output layer and multiple hidden layers.

The hidden layers in Transformer consist of three kinds of sublayers, including the self-attention, the cross-attention and the feed-forward network (FFN) sublayers. 
Both the self-attention and the cross-attention sublayers are based on the multi-head attention (MHA) mechanism. 
The multi-head attention takes the output of the previous sublayer as its input and computes a vector of attention scores (i.e., a distribution).
Then the FFN sublayer performs a non-linear transformation on the attention scores.

All sublayers are coupled with the residual connections \cite{DBLP:conf/cvpr/HeZRS16}, i.e., $Y=f(X)+X$, where $X$ denotes the layer input and $f$ could be any sublayer. 
Their inputs or outputs are also preprocessed by the layer normalization \cite{DBLP:journals/corr/BaKH16}.


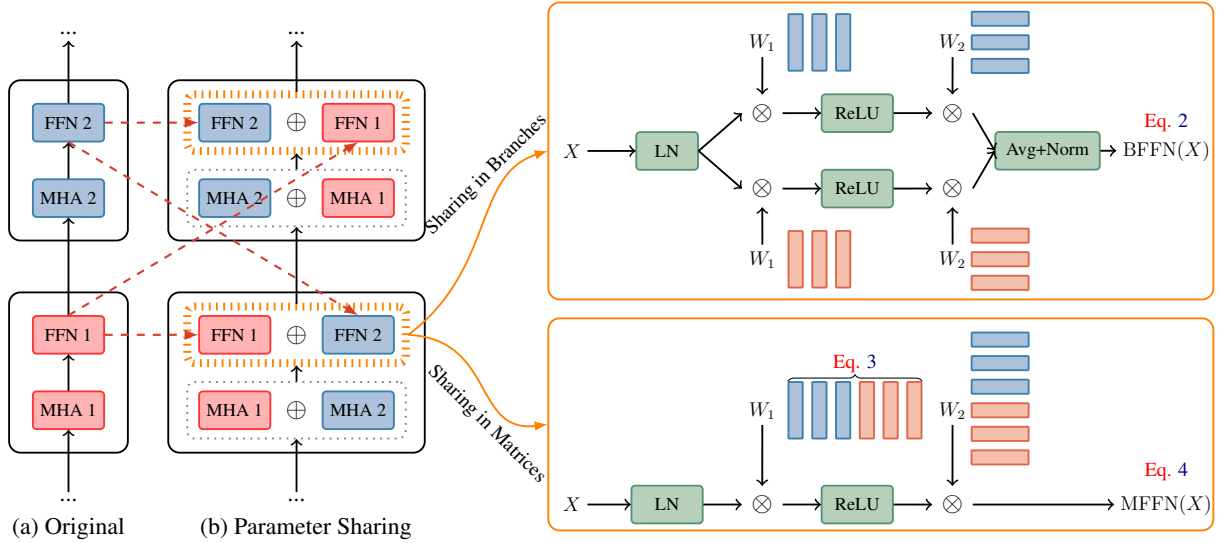
\begin{figure*}
  \centering
  \tikzstyle{layer} = [rectangle,line width = 0.7pt,rounded corners=0.05cm,minimum width=1.5cm,minimum height=0.8cm,inner sep=0.1cm]
  \tikzstyle{layer1} = [rectangle,line width = 0.7pt,rounded corners=0.05cm,minimum width=1.3cm,minimum height=0.8cm,inner sep=0.1cm]
  \tikzstyle{layer_small} = [rectangle,line width = 0.7pt,rounded corners=0.01cm,minimum width=0.3cm,minimum height=1.2cm,inner sep=0.1cm]
  \tikzstyle{layer_small_1} = [rectangle,line width = 0.7pt,rounded corners=0.01cm,minimum width=1.2cm,minimum height=0.3cm,inner sep=0.1cm]
  \tikzstyle{frame} = [rectangle,line width = 0.7pt,dotted,rounded corners=0.1cm,minimum width=4.6cm,minimum height=1.2cm, draw=gray]
  \tikzstyle{block} = [rectangle,line width = 0.7pt,rounded corners=0.15cm,minimum width=5.4cm,minimum height=3.4cm,draw=black!60]
  \tikzstyle{block1} = [rectangle,line width = 0.7pt,rounded corners=0.15cm,minimum width=2.5cm,minimum height=3.4cm,draw=black!60]
  \tikzstyle{conect} = [->,line width = 0.7pt]
  \begin{tikzpicture}[node distance = 0,scale = 0.625]
  \begin{scope}
  \tikzstyle{every node}=[scale=0.625]

  \begin{scope}
  \node(layer1)[layer,yshift=0.5cm,draw=red!80,fill=red!30]{};
  \node(text1)[above of = layer1]{MHA\ 1};
  \node(layer2)[layer,above of = layer1,yshift=1.6cm,draw=red!80,fill=red!30]{};
  \node(text2)[above of = layer2]{FFN\ 1};
  \node(block1_old)[block1,above of = layer1,yshift=0.8cm,draw=black]{};
  \node(point1_old)[below of = block1_old,yshift=-2.7cm,font=\Large]{...};
  \draw[conect](point1_old.north)to(layer1.south);
  \draw[conect](layer1.north)to(layer2.south);
  \node(layer3)[layer,above of = layer1, yshift=4.5cm,draw=lyyblue!80,fill=lyyblue!30]{};
  \node(text3)[above of = layer3]{MHA\ 2};
  \node(layer4)[layer,above of = layer3,yshift=1.6cm,draw=lyyblue!80,fill=lyyblue!30]{};
  \node(text4)[above of = layer4]{FFN\ 2};
  \node(block2_old)[block1,above of = layer3,yshift=0.8cm,draw=black]{};
  \node(point3_old)[above of = block2_old,yshift=2.7cm,font=\Large]{...};
  \draw[conect](layer4.north)to(point3_old.south);
  \draw[conect](layer2.north)to(layer3.south);
  \draw[conect](layer3.north)to(layer4.south);
  \end{scope}

  \begin{scope}
  \node(layer1_1)[layer,right of = layer1,xshift=3.5cm,draw=red!80,fill=red!30]{};
  \node(text1_1)[above of = layer1_1]{MHA\ 1};
  \node()[right of = layer1_1,xshift=1.3cm,scale=1.5]{$\oplus$};
  \node(layer1_2)[layer,right of = layer1_1,xshift=2.6cm,draw=lyyblue!80,fill=lyyblue!30]{MHA\ 2};
  \node(frame1)[frame,above of = layer1_1,xshift=1.3cm]{};
  \node(layer2_1)[layer,above of = layer1_1,yshift=1.6cm,draw=red!80,fill=red!30]{};
  \node(text2_1)[above of = layer2_1]{FFN\ 1};
  \node()[right of = layer2_1,xshift=1.3cm,scale=1.5]{$\oplus$};
  \node(layer2_2)[layer,above of = layer1_2,yshift=1.6cm,draw=lyyblue!80,fill=lyyblue!30]{FFN\ 2};
  \node(frame2)[frame,above of = layer2_1,xshift=1.3cm,line width = 3.0pt,draw=orange]{};
  \node(block1)[block,above of = layer1_1,xshift=1.3cm,yshift=0.8cm,draw=black]{};
  \node(point1)[below of = block1,yshift=-2.7cm,font=\Large]{...};
  \draw[conect](point1.north)to(frame1.south);
  \draw[conect](frame1.north)to(frame2.south);
  
  \node(layer3_1)[layer,above of = layer1_1,yshift=4.5cm,draw=lyyblue!80,fill=lyyblue!30]{};
  \node(text3_1)[above of = layer3_1]{MHA\ 2};
  \node()[right of = layer3_1,xshift=1.3cm,scale=1.5]{$\oplus$};
  \node(layer3_2)[layer,right of = layer3_1,xshift=2.6cm,draw=red!80,fill=red!30]{MHA\ 1};
  \node(frame3)[frame,above of = layer3_1,xshift=1.3cm]{};
  \node(layer4_1)[layer,above of = layer3_1,yshift=1.6cm,draw=lyyblue!80,fill=lyyblue!30]{};
  \node(text4_1)[above of = layer4_1]{FFN\ 2};
  \node()[right of = layer4_1,xshift=1.3cm,scale=1.5]{$\oplus$};
  \node(layer4_2)[layer,above of = layer3_2,yshift=1.6cm,draw=red!80,fill=red!30]{FFN\ 1};
  \node(frame4)[frame,above of = layer4_1,xshift=1.3cm,line width = 3.0pt,draw=orange]{};
  \node(block2)[block,above of = layer3_1,xshift=1.3cm,yshift=0.8cm,draw=black]{};
  \draw[conect](frame2.north)to(frame3.south);
  \draw[conect](frame3.north)to(frame4.south);
  \node(point3)[above of = block2,yshift=2.7cm,font=\Large]{...};
  \draw[conect](frame4.north)to(point3.south);
  \end{scope}

  \begin{scope}
  \node(X1)[right of = layer1_2,xshift=4.5cm,yshift=-2.0cm,font=\large]{$X$};
  \node(layer_ln1)[layer,right of = X1,xshift=2.0cm,draw=lyygreen!80,fill=lyygreen!30]{};
  \node(text_ln1)[above of = layer_ln1]{LN};
  \node(times11)[right of = text_ln1,xshift=2.0cm,scale=1.5]{$\otimes$};
  \node(layer_relu1)[layer,right of = times11,xshift=2.0cm,draw=lyygreen!80,fill=lyygreen!30]{};
  \node(text_relu1)[above of = layer_relu1]{ReLU};
  \node(times12)[right of = text_relu1,xshift=2.0cm,scale=1.5]{$\otimes$};
  \node(Y1)[right of = times12,xshift=4.5cm,font=\large]{$\mathrm{MFFN}(X)$};
  \draw[conect](X1.east)to(layer_ln1.west);
  \draw[conect](layer_ln1.east)to(times11.west);
  \draw[conect](times11.east)to(layer_relu1.west);
  \draw[conect](layer_relu1.east)to(times12.west);
  \draw[conect](times12.east)to(Y1.west);
  \node(W1)[above of = times11,yshift=2.0cm,font=\large]{$W_1$};
  \node(W2)[above of = times12,yshift=2.0cm,font=\large]{$W_2$};
  \draw[conect](W1.south)to(times11.north);
  \draw[conect](W2.south)to(times12.north);

  \node(cat1_begin)[layer_small,right of = W1,xshift=0.7cm,draw=lyyblue!80,fill=lyyblue!30]{};
  \node()[layer_small,right of = cat1_begin,xshift=0.5cm,draw=lyyblue!80,fill=lyyblue!30]{};
  \node()[layer_small,right of = cat1_begin,xshift=1.0cm,draw=lyyblue!80,fill=lyyblue!30]{};
  \node()[layer_small,right of = cat1_begin,xshift=1.5cm,draw=lyyred!80,fill=lyyred!30]{};
  \node()[layer_small,right of = cat1_begin,xshift=2.0cm,draw=lyyred!80,fill=lyyred!30]{};
  \node(cat1_end)[layer_small,right of = cat1_begin,xshift=2.5cm,draw=lyyred!80,fill=lyyred!30]{};
  \draw[decorate,decoration={brace}] (cat1_begin.north west) to node [above,yshift=0.05cm,font=\large,red] {\eqn{eqn:4}} (cat1_end.north east);

  \node(cat2_begin)[layer_small_1,right of = W2,xshift=1.0cm,yshift=-1.0cm,draw=lyyred!80,fill=lyyred!30]{};
  \node()[layer_small_1,above of = cat2_begin,yshift=0.5cm,draw=lyyred!80,fill=lyyred!30]{};
  \node()[layer_small_1,above of = cat2_begin,yshift=1.0cm,draw=lyyred!80,fill=lyyred!30]{};
  \node()[layer_small_1,above of = cat2_begin,yshift=1.5cm,draw=lyyblue!80,fill=lyyblue!30]{};
  \node()[layer_small_1,above of = cat2_begin,yshift=2.0cm,draw=lyyblue!80,fill=lyyblue!30]{};
  \node(cat2_end)[layer_small_1,above of = cat2_begin,yshift=2.5cm,draw=lyyblue!80,fill=lyyblue!30]{};

  \node(method_a)[above of = X1,xshift=6.5cm,yshift=1.7cm,rectangle,line width = 0.7pt,rounded corners=0.15cm,minimum width=14cm,minimum height=4.5cm,draw=black!60,line width = 0.7pt,draw=orange]{};
  \end{scope}

  \begin{scope}
  \node(X2)[above of = X1,yshift=7.5cm,font=\large]{$X$};
  \node(layer_ln2)[layer1,right of = X2,xshift=2.0cm,draw=lyygreen!80,fill=lyygreen!30]{};
  \node(text_ln2)[above of = layer_ln2]{LN};
  \node(times21)[right of = text_ln2,xshift=2.0cm,yshift=-0.8cm,scale=1.5]{$\otimes$};
  \node(layer_relu21)[layer,right of = times21,xshift=2.0cm,draw=lyygreen!80,fill=lyygreen!30]{};
  \node(text_relu21)[above of = layer_relu21]{ReLU};
  \node(times22)[right of = text_relu21,xshift=2.0cm,scale=1.5]{$\otimes$};
  \node(times23)[right of = text_ln2,xshift=2.0cm,yshift=0.8cm,scale=1.5]{$\otimes$};
  \node(layer_relu22)[layer,right of = times23,xshift=2.0cm,draw=lyygreen!80,fill=lyygreen!30]{};
  \node(text_relu22)[above of = layer_relu22]{ReLU};
  \node(times24)[right of = text_relu22,xshift=2.0cm,scale=1.5]{$\otimes$};
  \node(layer_norm)[layer,right of = times24,xshift=2.0cm,yshift=-0.8cm,minimum width=2.2cm,draw=lyygreen!80,fill=lyygreen!30]{};
  \node(text_norm)[above of = layer_norm]{Avg+Norm};
  \node(Y2)[right of = layer_norm,xshift=2.5cm,font=\large]{$\mathrm{BFFN}(X)$};
  \draw[conect](X2.east)to(layer_ln2.west);
  \draw[conect](layer_ln2.east)to(times21.west);
  \draw[conect](layer_ln2.east)to(times23.west);
  \draw[conect](times21.east)to(layer_relu21.west);
  \draw[conect](times23.east)to(layer_relu22.west);
  \draw[conect](layer_relu21.east)to(times22.west);
  \draw[conect](layer_relu22.east)to(times24.west);
  \draw[conect](times22.east)to(layer_norm.west);
  \draw[conect](times24.east)to(layer_norm.west);
  \draw[conect](layer_norm.east)to(Y2.west);
  
  \node(W1)[below of = times21,yshift=-1.5cm,font=\large]{$W_1$};
  \node(W2)[below of = times22,yshift=-1.5cm,font=\large]{$W_2$};
  \draw[conect](W1.north)to(times21.south);
  \draw[conect](W2.north)to(times22.south);
  \node(b11)[layer_small,right of = W1,xshift=0.7cm,draw=lyyred!80,fill=lyyred!30]{};
  \node()[layer_small,right of = b11,xshift=0.5cm,draw=lyyred!80,fill=lyyred!30]{};
  \node()[layer_small,right of = b11,xshift=1.0cm,draw=lyyred!80,fill=lyyred!30]{};
  \node(b12)[layer_small_1,right of = W2,xshift=1.0cm,yshift=-0.5cm,draw=lyyred!80,fill=lyyred!30]{};
  \node()[layer_small_1,above of = b12,yshift=0.5cm,draw=lyyred!80,fill=lyyred!30]{};
  \node()[layer_small_1,above of = b12,yshift=1.0cm,draw=lyyred!80,fill=lyyred!30]{};
  \node(W1_)[above of = times23,yshift=1.5cm,font=\large]{$W_1$};
  \node(W2_)[above of = times24,yshift=1.5cm,font=\large]{$W_2$};
  \draw[conect](W1_.south)to(times23.north);
  \draw[conect](W2_.south)to(times24.north);
  \node(b21)[layer_small,right of = W1_,xshift=0.7cm,draw=lyyblue!80,fill=lyyblue!30]{};
  \node()[layer_small,right of = b21,xshift=0.5cm,draw=lyyblue!80,fill=lyyblue!30]{};
  \node()[layer_small,right of = b21,xshift=1.0cm,draw=lyyblue!80,fill=lyyblue!30]{};
  \node(b22)[layer_small_1,right of = W2_,xshift=1.0cm,yshift=-0.5cm,draw=lyyblue!80,fill=lyyblue!30]{};
  \node()[layer_small_1,above of = b22,yshift=0.5cm,draw=lyyblue!80,fill=lyyblue!30]{};
  \node()[layer_small_1,above of = b22,yshift=1.0cm,draw=lyyblue!80,fill=lyyblue!30]{};
  \node(eqn3)[above of = layer_norm,xshift=2.5cm,yshift=0.6cm,font=\large,red]{\eqn{eqn:3}};
  \node(eqn5)[below of = eqn3,xshift=0.0cm,yshift=-7.4cm,font=\large,red]{\eqn{eqn:5}};

  \node(method_b)[above of = X2,xshift=6.5cm,yshift=0.0cm,rectangle,line width = 0.7pt,rounded corners=0.15cm,minimum width=14cm,minimum height=6.3cm,draw=black!60,line width = 0.7pt,draw=orange]{};
  \end{scope}

  \node()[above of = layer1,yshift=-2.5cm,font=\Large]{(a) Original};
  \node()[above of = layer1_1,xshift=1.5cm,yshift=-2.5cm,font=\Large]{(b) Parameter Sharing};

  \draw[-latex,line width = 0.7pt,draw=orange] (frame2.east)..controls (9.0,2.0)and(8.5,0.0) .. (method_a.west);
  \draw[-latex,line width = 0.7pt,draw=orange] (frame2.east)..controls (9.0,3.5)and(8.5,5.5) .. (method_b.west);

  \node()[above of = method_a,xshift=-8.3cm,yshift=0.0cm,font=\large,rotate=-45]{Sharing in Matrices};
  \node()[above of = method_a,xshift=-8.3cm,yshift=5.3cm,font=\large,rotate=45]{Sharing in Branches};

  \draw[-latex,dashed,thick,lyyred] (layer2.east) to (layer2_1.west);
  \draw[-latex,dashed,thick,lyyred] (layer4.east) to (layer4_1.west);
  \draw[-latex,dashed,thick,lyyred] (layer4.south) to (layer2_2.north);
  \draw[-latex,dashed,thick,lyyred] (layer2.north) to (layer4_2.south);

  \end{scope}
  \end{tikzpicture}
  \label{}
  \caption{Comparison of sharing parameters in branches and matrices (A running example of the feed-forward network. Different colors denote parameters from different sources. For simplicity, we omit $b_1$ and $b_2$ in \eqn{eqn:ffn}. ).}
  \label{fig:comparison1}
\end{figure*}

\subsection{Parameters Sharing}
\label{sub-sec:hypothesis}
Although the large Transformer model has shown its effectiveness \cite{DBLP:conf/nips/VaswaniSPUJGKP17,DBLP:conf/iclr/BaevskiA19,DBLP:conf/nips/BrownMRSKDNSSAA20}, only models with very limited parameters can be deployed on resource-limited devices \cite{DBLP:journals/corr/HanMD15,DBLP:journals/pieee/DengLHSX20,DBLP:journals/corr/abs-2002-11794}.
As a simple and effective approach, parameter sharing can improve the model performance under the limited parameters \cite{DBLP:conf/iclr/DehghaniGVUK19,DBLP:journals/corr/abs-2108-10417,DBLP:journals/corr/abs-2104-06022,DBLP:conf/iclr/LanCGGSS20}.  
Most research in Transformers focuses on sharing parameters along model depth, and models with parameter sharing will generally achieve better performance than those without sharing.
Here we further study this approach and hypothesize its success may come from two perspectives:
\begin{itemize}
  \item Model complexity (Section \ref{sec:complexity}).
  \item Training convergence (Section \ref{sec:convergence}).
\end{itemize}

\section{Model Complexity}
\label{sec:complexity}

To explore the impact of increasing model complexity, in this section, we first propose increasing model complexity by sharing parameters in two other dimensions: model branches and weight matrices.
Then, we analyze the results of sharing parameters and draw the preliminary conclusion that:

\uwave{Model complexity does have a positive impact on performance in parameter sharing.}




\subsection{Model Structure}
For sharing parameters in branches, we adopt a sublayer-level multi-branch network \cite{DBLP:journals/corr/abs-1711-02132,DBLP:journals/corr/abs-2006-10270,DBLP:conf/iclr/ShazeerMMDLHD17,DBLP:conf/icml/SoLL19,DBLP:journals/corr/abs-1911-09483}.
We take the feed-forward sublayer for an example in \fig{fig:comparison1}. 
The original FFN applies non-linear transformation to its input $X$ and can be denoted as:
\begin{equation}
    \mathrm{FFN}(X)=\mathrm{ReLU}(XW_1+b_1)W_2+b_2\label{eqn:ffn}
\end{equation}
where $W_1\in\mathbb{R}^{d\times 4d}$, $b_1\in\mathbb{R}^{4d}$, $W_2\in\mathbb{R}^{4d\times d}$ and $b_2\in\mathbb{R}^{d}$.

Suppose there is a 2-layer Transformer, we share parameters from these two sublayer to the two branches in every layer (denoted by the dashed red lines with arrows in \fig{fig:comparison1}), and every parameter from the original model will be reused twice.
To combine information from different branches, here we adopt a simple average operation.
For regularization and ease of training, we place an additional normalization unit on each average operation.
Such a newly constructed multi-branch feed-forward sublayer $\mathrm{BFFN}$ can be denoted as:
\begin{equation}
  \mathrm{BFFN}(X)=\mathrm{Norm}(\frac{\sum_{i=1}^n{\mathrm{FFN}}_i(X)}{n})
  \label{eqn:3}
\end{equation}
where $X$ and $Y$ are the input and output of the current sublayer, $n$ is the number of branches, and $\mathrm{FFN}_i(\cdot)$ denotes the $i$-th branch of the newly constructed feed-forward sublayer.

For sharing parameters in weight matrices, we concatenate parameters from different layers together to form a weight matrix with larger dimension.
In the lower right corner of \fig{fig:comparison1}, we take $W_1$ in \eqn{eqn:ffn} for example.
Let $W_1^{l_1}\in \mathbb{R}^{d\times 4d},...,W_1^{l_n}\in \mathbb{R}^{d\times 4d}$ denote weights from different layers, then the newly generated weight matrix $W_1^{cat}\in \mathbb{R}^{d\times 8d}$ can be rewritten as:
\begin{equation}
  W_1^{cat}=\mathrm{Concat}(W_1^{l_1},...,W_1^{l_n})
  \label{eqn:4}
\end{equation}
$W_2$, $b_1$, and $b_2$ are computed in the same way as $W_1$.
The newly constructed sublayer with larger matrices $\mathrm{MFFN}$ can be denoted as:
\begin{equation}
    \mathrm{MFFN}(X)=\mathrm{ReLU}(XW_1^{cat}+b_1^{cat})W_2^{cat}+b_2^{cat}
    \label{eqn:5}
\end{equation}

\begin{table}[t!]
  \centering
  \small
  \renewcommand\arraystretch{1.1}
  \setlength{\tabcolsep}{1.5mm}
  \begin{tabular}{llll}
  \hline
  \multicolumn{1}{l}{Model Structure} &
  \multicolumn{1}{l}{Params} &
  \multicolumn{1}{l}{Complexity} &
  \multicolumn{1}{l}{Parallelism} \\
  \hline
  \multirow{1}{*}{Original} & $O(d)$ & $O(d)$ & $O(1/d)$ \\
  \multirow{1}{*}{Sharing in layers} & $O(d)$ & $O(nd)$ & $O(1/nd)$ \\
  \multirow{1}{*}{Sharing in branches} & $O(d)$ & $O(nd)$ & $O(1/d)$ \\
  \multirow{1}{*}{Sharing in matrices} & $O(d)$ & $O(nd)$ & $O(1/d)$ \\
  \hline
  \end{tabular}
  \caption{The parameters, model complexity (FLOPs) and computational parallelism of different models. $d$ is the depth of the original model. $n$ means that each parameter is shared $n$ times, $n>1$.}
  \label{tab:complexity}
\end{table}

\begin{table*}[t!]
  \centering
  \setlength{\tabcolsep}{0.08cm}
  \small
  \renewcommand\arraystretch{1.0}
  \begin{tabular}{l|l|ccc|ccc|c|c|c}
  \hline
  \multirow{2}{*}{System} &
  \multirow{2}{*}{Depths} &
  \multicolumn{3}{c|}{WMT14 En-De} &
  \multicolumn{3}{c|}{WMT14 En-Fr} & \multirow{2}{*}{FLOPs} & \multirow{2}{*}{Steps} & \multirow{2}{*}{Costs} \\
  \cline{3-8}
  && Params & Test & Valid &
  Params & Test & Valid & & \\
  \hline
  \multirow{1}{*}{Transformer \cite{DBLP:conf/nips/VaswaniSPUJGKP17}} & 6-6 & 213M & 28.40 & - & 222M & 41.00 & - & 6.27G & 300K & 40.00 \\
  \multirow{1}{*}{ADMIN \cite{DBLP:journals/corr/abs-2008-07772}} & 60-12 & 262M & 30.01 & - & - & 43.80 & - & - & - \\
  \multirow{1}{*}{Depth Growing \cite{DBLP:conf/acl/WuWXTGQLL19}} & 8-8 & 270M & 29.92 & - & - & 43.27 & - & - & - \\
  \multirow{1}{*}{MUSE \cite{DBLP:journals/corr/abs-1911-09483}} & 12-12 & - & 29.90 & - & - & 43.50 & - & - & - \\
  \multirow{1}{*}{MAT \cite{DBLP:journals/corr/abs-2006-10270}} & 6-6 & 206M & 29.90 & - & - & - & - & - & - \\
  \multirow{1}{*}{Evolved Transformer \cite{DBLP:conf/icml/SoLL19}} & 6-6 & 222M & 29.30 & - & 221M & 41.30 & - & - & - \\
  \multirow{1}{*}{Scaling NMT \cite{DBLP:conf/wmt/OttEGA18}} & 6-6 & 210M & 29.30 & - & 222M & 43.20 & - & - & - \\
  \multirow{1}{*}{DLCL \cite{DBLP:conf/acl/WangLXZLWC19}} & 30-6 & 137M & 29.30 & - & - & - & - & - & - \\
  \multirow{1}{*}{Weighted \cite{DBLP:journals/corr/abs-1711-02132}} & 6-6 & 213M & 28.90 & - & - & 41.40 & - & - & - \\
  \multirow{1}{*}{Universal \cite{DBLP:conf/iclr/DehghaniGVUK19}} & - & - & 28.90 & - & - & - & - & - & - \\
  \hline
  \multirow{1}{*}{Transformer-base} & 6-6 & 63M & 27.56 & 25.94 & 111M & 40.72 & 46.80 & 1.81G & 100K & 8.68 \\
  \multirow{1}{*}{+ SIL \cite{DBLP:journals/corr/abs-2104-06022}} & 24-6 & 63M & 28.27 & \colorbox{gray!10}{\underline{\textbf{26.70}}} & 112M & 41.84 & 47.81 & 3.51G & 100K & 17.08 \\
  \multirow{1}{*}{+ SIB (discussed in Section \ref{sec:complexity})} & 6-6 & 63M & 28.32 & 26.50 & 112M & 41.71 & 47.83 & 3.51G & 100K & 18.20 \\
  \multirow{1}{*}{+ SIM (discussed in Section \ref{sec:complexity})} & 6-6 & 63M & 28.32 & \colorbox{gray!10}{\underline{\textbf{26.70}}} & 112M & 41.64 & 47.66 & 3.51G & 100K & 12.88 \\
  \multirow{1}{*}{+ Tuning Hyper. (discussed in Section \ref{sec:convergence})} & 6-6 & 63M & \colorbox{gray!10}{\underline{\textbf{28.62}}} & 26.67 & 112M & \colorbox{gray!10}{\underline{\textbf{42.01}}} & \colorbox{gray!10}{\underline{\textbf{47.92}}} & 1.81G & 100K & 13.33 \\
  \hline
  \multirow{1}{*}{Transformer-deep-12l} & 12-6 & 83M & 28.45 & 26.82 & 111M & 41.48 & 47.88 & 2.38G & 100K & 11.76 \\
  \multirow{1}{*}{+ SIL \cite{DBLP:journals/corr/abs-2104-06022}} & 48-6 & 83M & 29.13 & \colorbox{gray!10}{\underline{\textbf{27.11}}} & 112M & 42.31 & 48.45 & 5.78G & 100K & 28.28 \\
  \multirow{1}{*}{+ SIB (discussed in Section \ref{sec:complexity})} & 12-6 & 83M & 28.81 & 26.90 & 112M & 41.94 & \colorbox{gray!10}{\underline{\textbf{48.57}}} & 5.78G & 100K & 29.96 \\
  \multirow{1}{*}{+ SIM (discussed in Section \ref{sec:complexity})} & 12-6 & 83M & 28.86 & 26.92 & 112M & 42.38 & 48.48 & 5.78G & 100K & 19.32 \\
  \multirow{1}{*}{+ Tuning Hyper. (discussed in Section \ref{sec:convergence})} & 12-6 & 83M & \colorbox{gray!10}{\underline{\textbf{29.49}}} & 26.97 & 112M & \colorbox{gray!10}{\underline{\textbf{42.40}}} & 48.55 & 2.38G & 100K & 17.50 \\
  \hline
  \end{tabular}
  \caption{Results of different systems on WMT14 En-De and WMT14 En-Fr (SIL: Sharing parameters in different layers; SIB: Sharing parameters in different branches; SIM: Sharing parameters in different weight matrices; Costs: Training costs that are normalized to GPU hours on a single NVIDIA Tesla V100 GPU on WMT14 En-De;  FLOPs is measured on a sample with src/tgt length of 30 and 32K vocabulary.).}
  \label{tab:main_result_ende}
\end{table*}

\subsection{Comparison of Different Models}
To better understand the advantages and disadvantages of different models in \tab{tab:complexity}, here we compare the original model with three parameter sharing models.
We take the reciprocal value of the model depth to measure the computational parallelism of each model.
As shown in the last three lines of \tab{tab:complexity}, although the parameter sharing models have the same number of parameters as the original model, the model complexity will increase proportionally with shared times $n$.
Compared to sharing parameters in layers, sharing parameters in branches and matrices can provide higher computational parallelism ($O(1/d)>O(1/nd)$) to be efficiently computed on modern GPUs, which can be also reflected in our experimental results.

\begin{table}[!t]
  \centering
  \small
  \renewcommand\arraystretch{1.1}
  \setlength{\tabcolsep}{1.5mm}{
  \begin{tabular}{c|r|r|r|r|r|r}
  \hline
  \multicolumn{1}{c|}{\multirow{2}*{Lang.}}&
  \multicolumn{2}{c|}{Train} &
  \multicolumn{2}{c|}{Test} &
  \multicolumn{2}{c}{Valid}\\
  \cline{2-7}
  & \multicolumn{1}{c|}{Sent.} & \multicolumn{1}{c|}{Word} & \multicolumn{1}{c|}{Sent.} & \multicolumn{1}{c|}{Word} & \multicolumn{1}{c|}{Sent.} & \multicolumn{1}{c}{Word} \\
  \cline{1-7}
  \href{http://statmt.org/wmt14/translation-task.html}{En-De} & 4.5M & 262M & 3003 & 164K & 3000 & 156K \\
  \cline{1-7}
  \href{http://statmt.org/wmt14/translation-task.html}{En-Fr} & 35M & 2.4G & 3003 & 175K & 26K & 1.8M \\
  \hline
  \end{tabular}
  \caption{Date statistics (\# of sentences and \# of words).}
  \label{tab:data1}
  }
\end{table}


\subsection{Experiments in Section \ref{sec:complexity}}
\label{sub-sec:experiments-section3}

We evaluate the experiments in Section \ref{sec:complexity} on WMT14 English-German (En-De) and English-French (En-Fr)  machine translation tasks. 
Detailed data statistics as well as the URLs are shown in \tab{tab:data1} and Appendix \ref{sec:appendixa}.
We experiment with the Transformer-base and Transformer-deep settings in \tab{tab:main_result_ende}. 
Experiments on Transformer-big are shown in \tab{tab:main_result_big}.
The standard Transformer-base and Transformer-big systems both consist of a 6-layer encoder and a 6-layer decoder.
The Transformer-deep system has 12 layers.
We also adopt the relative positional representation (RPR) \cite{DBLP:conf/naacl/ShawUV18} to construct stronger baselines.
All experiments are tested on the model ensemble by averaging the last 5 checkpoints.

\begin{table*}[t!]
  \centering
  \setlength{\tabcolsep}{0.05cm}
  \small
  \renewcommand\arraystretch{1.0}
  \begin{tabular}{l|l|ccc|ccc|c|c|c}
  \hline
  \multirow{2}{*}{System} &
  \multirow{2}{*}{Depths} &
  \multicolumn{3}{c|}{WMT14 En-De} &
  \multicolumn{3}{c|}{WMT14 En-Fr} & \multirow{2}{*}{FLOPs} & \multirow{2}{*}{Steps} & \multirow{2}{*}{Costs} \\
  \cline{3-8}
  && Params & Test & Valid & Params & Test & Valid &
  & & \\
  \hline
  \multirow{1}{*}{Transformer-big} & 6-6 & 213M & 28.98 & 26.92 & 311M & 41.95 & 48.26 & 6.27G & 300K & 40.00 \\
  \hline
  \multirow{1}{*}{+ SIL \cite{DBLP:journals/corr/abs-2104-06022}} & 24-6 & 213M & \multicolumn{2}{c|}{\sout{loss exploding} \ding{55}} & 316M & 43.22 & \colorbox{gray!10}{\underline{\textbf{49.91}}} & 13.06G & - & - \\
  \multirow{1}{*}{\quad + $L_2$ Regularization} & 24-6 & 213M & 29.70 & 27.14 & 316M & \colorbox{gray!10}{\underline{\textbf{43.23}}} & 49.76 & 13.06G & 300K & 80.00 \\
  \hline
  \multirow{1}{*}{+ SIB (discussed in Section \ref{sec:complexity})} & 6-6 & 213M & \multicolumn{2}{c|}{\sout{loss exploding} \ding{55}} & 316M & 42.98 & 49.81 & 13.06G & - & - \\
  \multirow{1}{*}{\quad + $L_2$ Regularization} & 6-6 & 213M & 29.50 & 26.94 & 316M & 43.09 & 49.84 & 13.06G & 300K & 83.33 \\
  \hline
  \multirow{1}{*}{+ SIM (discussed in Section \ref{sec:complexity})} & 6-6 & 213M & \multicolumn{2}{c|}{\sout{loss exploding} \ding{55}} & 316M & \multicolumn{2}{c|}{\sout{loss exploding} \ding{55}} & 13.06G & - & - \\
  \multirow{1}{*}{\quad + $L_2$ Regularization} & 6-6 & 213M & \colorbox{gray!10}{\underline{\textbf{29.98}}} & 26.94 & 316M & 43.15 & 49.68 & 13.06G & 300K & 70.00 \\
  \hline
  \multirow{1}{*}{+ Tuning Hyper. (discussed in Section \ref{sec:convergence})} & 6-6 & 213M & \multicolumn{2}{c|}{\sout{loss exploding} \ding{55}} & 316M & \multicolumn{2}{c|}{\sout{loss exploding} \ding{55}} & 6.27G & - & - \\
  \multirow{1}{*}{\quad + $L_2$ Regularization} & 6-6 & 213M & 29.87 & \colorbox{gray!10}{\underline{\textbf{27.15}}} & 316M & 42.88 & 49.62 & 6.27G & 240K & 58.67 \\
  \hline
  \end{tabular}
  \caption{Results of Transformer-big systems on WMT14 En-De and WMT14 En-Fr.}
  \label{tab:main_result_big}
\end{table*}

\textbf{All parameter sharing approaches are parameter-efficient.}
\tab{tab:main_result_ende} shows the results of various Transformer benchmarks and these three sharing strategies. 
SIL, SIB and SIM all have significantly higher parameter efficiency compared to several state-of-the-art systems.
For example, compared to the Transformer with 213M parameters (28.40), SIM obtains a similar performance (28.32) with less than 1/3 parameters (63M) on WMT14 En-De test data.
With the base configuration, SIL/SIB/SIM achieves 28.27/28.32/28.32 BLEU points on En-De, which is 0.71/0.76/0.76 higher than the standard Transformer-base.
Experiments on Transformer-deep and Transformer-big also show similar results.

\textbf{Sharing in matrices has the highest training efficiency.}
Among these three sharing strategies, SIM has higher computational parallelism and shows the highest training efficiency.
Compared to SIL, SIM can save 24.59\%/31.68\% of training time on Transformer-base/12l.
As for SIB, although the multi-branch network has an inherent advantage of high computational parallelism as discussed in \tab{tab:complexity}, due to the limitations of practical computational libraries, the computational efficiency of this structure does not reach its theoretical height.

\textbf{Model complexity does have a positive impact.}
As shown in \tab{tab:main_result_ende} and \tab{tab:main_result_big}, SIL, SIB, and SIM show better performance than the original model in several different settings.
It reflects that, under a constant number of parameters, increasing model complexity does have a positive impact on model performance.
However, due to the impact of parameter sharing on model convergence, we need further experiments to verify whether model complexity is the decisive factor.

\begin{figure}
  \centering
  \small
  \begin{tikzpicture}
      \begin{axis}[
          width=1.00\linewidth,height=0.48\linewidth,
          yticklabel style={/pgf/number format/fixed,/pgf/number format/precision=1},
          ylabel={Loss},
          ylabel near ticks,
          xlabel={\#Epoch (4,800 steps/epoch)},
          xlabel near ticks,
          enlargelimits=0.05,
          xtick distance=2,
          xmajorgrids=true,
          ymajorgrids=true,
          grid style=dashed,
          every tick label/.append style={font=\small},
          label style={font=\small},
          ylabel style={yshift=5pt},
          legend style={font=\small,inner sep=3pt},
          legend image post style={scale=1},
          legend columns=2,
          legend cell align={left},
      ]
      \addplot [orange,thick,mark=*] coordinates {
        (2,4.545) (4,4.234) (6,4.124) (8,4.075) (10,4.065) (12,4.046) (14,4.025) (16,4.010)  (18,4.008) (20,4.003) (21,3.994)
      };
      \addlegendentry{Original}
      
      \addplot [lyygreen,thick,mark=square*] coordinates {
          (2,4.492) (4,4.172) (6,4.073) (8,4.022) (10,4.010) (12,3.994) (14,3.983) (16,3.975) (18,3.960) (20,3.962) (21,3.952)
      };
      \addlegendentry{SIB}
      \addplot [lyyblue,thick,mark=diamond*] coordinates {
          (2,4.525) (4,4.181) (6,4.083) (8,4.040) (10,4.023) (12,4.003) (14,3.991) (16,3.977) (18,3.979) (20,3.970) (21,3.956)
      };
      \addlegendentry{SIL}
      \addplot [lyyred,thick,mark=triangle*] coordinates {
          (2,4.512) (4,4.170) (6,4.071) (8,4.026) (10,4.013) (12,3.995) (14,3.982) (16,3.959) (18,3.957) (20,3.960) (21,3.948)
      };
      \addlegendentry{SIM}
      \end{axis}
  \end{tikzpicture}
  \caption{Loss vs. the number of epochs on En-De (learning rate:$0.001$; warmup:8,000; batch:4,096).}
  \label{fig:curve1}
\end{figure}
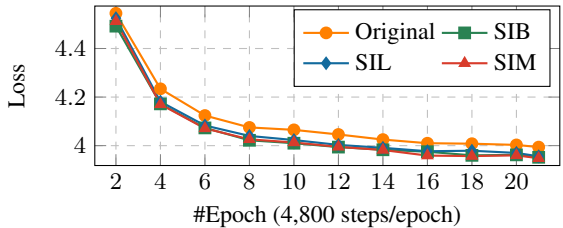

\section{Training Convergence}
\label{sec:convergence}

This section focuses on the relationship between parameter sharing, hyperparameter tuning, and training convergence.
Experimental results show that:

\uwave{The success of parameter sharing on performance depends largely on training convergence ($\Delta\mathrm{BLEU}\approx1.0$) and a small part on model complexity ($\Delta\mathrm{BLEU}<0.5$).}


\subsection{Parameter Sharing vs. Convergence}
In the parameter sharing models, since each parameter is shared multiple times, one parameter will participate in the network computation several times in a single training step. 
Its corresponding gradient will accumulate larger in back propagation, so the convergence speed of the parameter sharing model is always faster than the original model under the same learning rate schedule.
To verify this, we plot validation loss curves for the original baseline, SIL, SIB, and SIM to observe their convergence.
As shown in \fig{fig:curve1}, SIL/SIB/SIM always has a lower loss than the baseline model, and the lower loss does reflect better performance in our experiments.

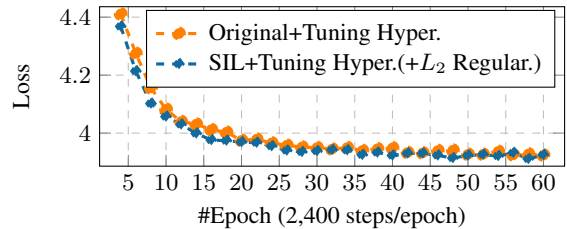
\begin{figure}
  \centering
  \small
  \begin{tikzpicture}
      \begin{axis}[
          width=1.00\linewidth,height=0.48\linewidth,
          yticklabel style={/pgf/number format/fixed,/pgf/number format/precision=1},
          ylabel={Loss},
          ylabel near ticks,
          xlabel={\#Epoch (2,400 steps/epoch)},
          xlabel near ticks,
          enlargelimits=0.05,
          xtick distance=5,
          xmajorgrids=true,
          ymajorgrids=true,
          grid style=dashed,
          every tick label/.append style={font=\small},
          label style={font=\small},
          ylabel style={yshift=5pt},
          legend style={font=\small,inner sep=3pt},
          legend image post style={scale=1},
          legend columns=1,
          legend cell align={left},
      ]
      \addplot [orange,very thick,dashed,mark=*] coordinates {
        (4,4.411) (6,4.275) (8,4.156) (10,4.082) (12,4.040) (14,4.031) (16,4.012)  (18,4.003) (20,3.975) (22,3.979) (24,3.966) (26,3.958) (28,3.952) (30,3.950) (32,3.945) (34,3.950) (36,3.940) (38,3.946) (40,3.949) (42,3.933) (44,3.932) (46,3.940) (48,3.941) (50,3.927) (52,3.927) (54,3.937) (56,3.925) (58,3.928) (60,3.924)
      };
      \addlegendentry{Original+Tuning Hyper.}
      \addplot [lyyblue,very thick,dashed,mark=diamond*] coordinates {
        (4,4.368) (6,4.214) (8,4.102) (10,4.057) (12,4.030) (14,3.999) (16,3.976)  (18,3.974) (20,3.969) (22,3.967) (24,3.955) (26,3.940) (28,3.935) (30,3.938) (32,3.942) (34,3.941) (36,3.925) (38,3.933) (40,3.923) (42,3.930) (44,3.930) (46,3.923) (48,3.914) (50,3.923) (52,3.926) (54,3.921) (56,3.932) (58,3.911) (60,3.925)
      };
      \addlegendentry{SIL+Tuning Hyper.(+$L_2$ Regular.)}
      \end{axis}
  \end{tikzpicture}
  \caption{Loss vs. the number of epochs on En-De (learning rate:$0.002$; warmup:16,000; batch:8,192).}
  \label{fig:curve2}
\end{figure}

\begin{table}[!t]
  \centering
  \small
  \renewcommand\arraystretch{1.0}
  \setlength{\tabcolsep}{1.2mm}{
  \begin{tabular}{l|l|ccc}
  \hline
  \multicolumn{1}{c|}{\multirow{1}*{System}}&
  \multicolumn{1}{c|}{Depths} &
  \multicolumn{1}{c}{Params} &
  \multicolumn{1}{c}{Test} &
  \multicolumn{1}{c}{Valid} \\
  \hline
  Transformer-base & 6-6 & 63M & 27.56 & 25.94 \\
  \hline
  + Tuning Hyper. & 6-6 & 63M & 28.62 & 26.67 \\
  \hline
  + SIL & 24-6 & 63M & 28.27 & 26.70 \\
  \quad + Tuning Hyper. & 24-6 & 63M & \multicolumn{2}{c}{\sout{loss exploding} \ding{55}} \\
  \quad \quad + $L_2$ Regular. & 24-6 & 63M & \colorbox{gray!10}{\underline{\textbf{28.69}}} & \colorbox{gray!10}{\underline{\textbf{26.75}}} \\
  \hline
  \end{tabular}
  \caption{Results of systems in \fig{fig:curve1} and \fig{fig:curve2}.}
  \label{tab:tune_SIL}
  }
\end{table}

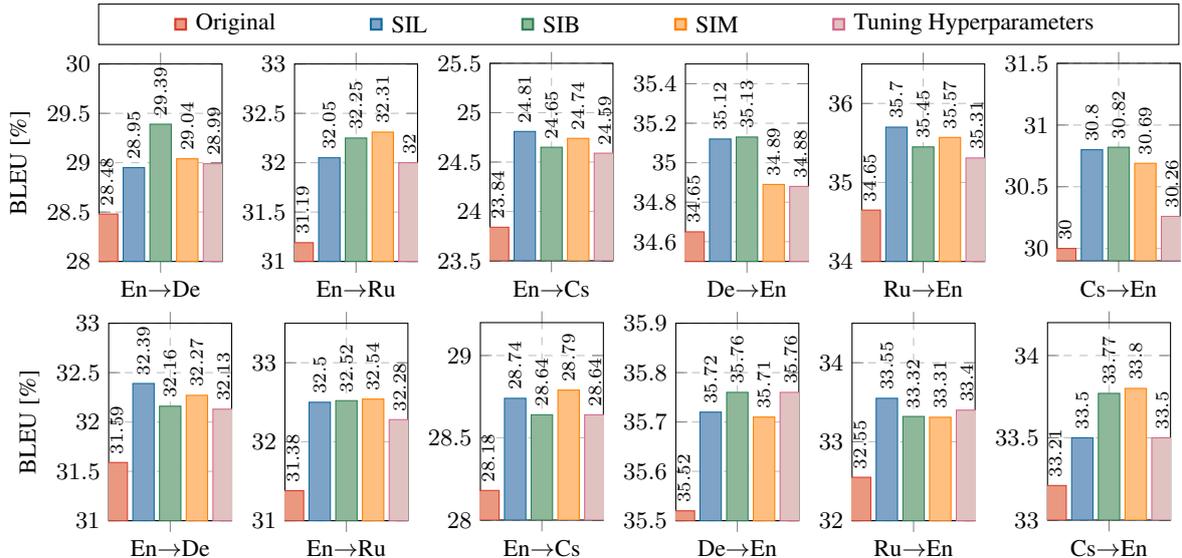
\begin{figure*}[t]
  \hspace{1.23cm}
  \tikz {
      \small
      \legendmargin=0.062\linewidth
      \legendwidth=0.01\linewidth,
      \legendsep=0.07\linewidth
      \coordinate (start) at (0,0);
      \draw[draw=lyyred,fill=lyyred!50,thick,postaction={decorate},decoration={markings,mark=at position 0.5 with {\pgfuseplotmark{square*}}}] ([xshift=\legendmargin]start.east) -- +(\legendwidth,0) node[black,right] (l1) {Original};
      \draw[draw=lyyblue,fill=lyyblue!50,thick,postaction={decorate},decoration={markings,mark=at position 0.5 with {\pgfuseplotmark{square*}}}] ([xshift=\legendsep]l1.east) -- +(\legendwidth,0) node[black,right] (l2) {SIL};
      \draw[draw=lyygreen,fill=lyygreen!50,thick,postaction={decorate},decoration={markings,mark=at position 0.5 with {\pgfuseplotmark{square*}}}] ([xshift=\legendsep]l2.east) -- +(\legendwidth,0) node[black,right] (l3) {SIB};
      \draw[draw=orange,fill=orange!50,thick,postaction={decorate},decoration={markings,mark=at position 0.5 with {\pgfuseplotmark{square*}}}] ([xshift=\legendsep]l3.east) -- +(\legendwidth,0) node[black,right] (l4) {SIM};
      \draw[draw=purple!60,fill=lypurple,thick,postaction={decorate},decoration={markings,mark=at position 0.5 with {\pgfuseplotmark{square*}}}] ([xshift=\legendsep]l4.east) -- +(\legendwidth,0) node[black,right] (l5) {Tuning Hyperparameters};
      \coordinate (end) at ([xshift=\legendmargin+0pt]l5.east);
      \begin{pgfonlayer}{background}
      \node[rectangle,draw,inner sep=0.2pt] [fit = (start) (l1) (l2) (l3) (l4) (l5) (end)] {};
      \end{pgfonlayer}
  }
  \\
  \centering
  \small
  \begin{tikzpicture}  
    \begin{axis}  
    [  
      width=3.2cm, height=4.2cm,
      ybar, 
      enlarge x limits=0.5,
      bar width=8pt,
      grid style=dashed,
      ymajorgrids=true,
      xmajorgrids=true,
      ylabel near ticks,
      xlabel near ticks,
      ylabel={BLEU [\%]},
      symbolic x coords={En$\rightarrow$De},
      xtick=data,  
      nodes near coords,
      ymin=28.0,
      ymax=30.0,
      every node near coord/.append style={
        font=\scriptsize,
        rotate=90, 
        anchor=west, 
      }
      ]  
    \addplot[draw=lyyred,fill=lyyred!50] coordinates {(En$\rightarrow$De, 28.48)};
    \addplot[draw=lyyblue,fill=lyyblue!50] coordinates {(En$\rightarrow$De, 28.95)};
    \addplot[draw=lyygreen,fill=lyygreen!50] coordinates {(En$\rightarrow$De, 29.39)};
    \addplot[draw=orange,fill=orange!50] coordinates {(En$\rightarrow$De, 29.04)};
    \addplot[draw=purple!60,fill=lypurple] coordinates {(En$\rightarrow$De, 28.99)};
    \end{axis}  
  \end{tikzpicture}  
  \begin{tikzpicture}  
    \begin{axis}  
    [  
      width=3.2cm, height=4.2cm,
      ybar, 
      enlarge x limits=0.5,
      bar width=8pt,
      grid style=dashed,
      ymajorgrids=true,
      xmajorgrids=true,
      ylabel near ticks,
      xlabel near ticks,
      symbolic x coords={En$\rightarrow$Ru},
      xtick=data,  
      nodes near coords,
      ymin=31.0, ymax=33.0,
      every node near coord/.append style={
        font=\scriptsize,
        rotate=90, 
        anchor=west, 
      }
      ]  
    \addplot[draw=lyyred,fill=lyyred!50] coordinates {(En$\rightarrow$Ru, 31.19)};
    \addplot[draw=lyyblue,fill=lyyblue!50] coordinates {(En$\rightarrow$Ru, 32.05)};
    \addplot[draw=lyygreen,fill=lyygreen!50] coordinates {(En$\rightarrow$Ru, 32.25)};
    \addplot[draw=orange,fill=orange!50] coordinates {(En$\rightarrow$Ru, 32.31)};
    \addplot[draw=purple!60,fill=lypurple] coordinates {(En$\rightarrow$Ru, 32.00)};
    \end{axis}  
  \end{tikzpicture}  
  \begin{tikzpicture}  
    \begin{axis}  
    [  
      width=3.2cm, height=4.2cm,
      ybar, 
      enlarge x limits=0.5,
      bar width=8pt,
      grid style=dashed,
      ymajorgrids=true,
      xmajorgrids=true,
      ylabel near ticks,
      xlabel near ticks,
      symbolic x coords={En$\rightarrow$Cs},
      xtick=data,  
      nodes near coords,
      ymin=23.5, ymax=25.5,
      every node near coord/.append style={
        font=\scriptsize,
        rotate=90, 
        anchor=west, 
      }
      ]  
    \addplot[draw=lyyred,fill=lyyred!50] coordinates {(En$\rightarrow$Cs, 23.84)};
    \addplot[draw=lyyblue,fill=lyyblue!50] coordinates {(En$\rightarrow$Cs, 24.81)};
    \addplot[draw=lyygreen,fill=lyygreen!50] coordinates {(En$\rightarrow$Cs, 24.65)};
    \addplot[draw=orange,fill=orange!50] coordinates {(En$\rightarrow$Cs, 24.74)};
    \addplot[draw=purple!60,fill=lypurple] coordinates {(En$\rightarrow$Cs, 24.59)};
    \end{axis}  
  \end{tikzpicture}  
  \begin{tikzpicture}  
    \begin{axis}  
    [  
      width=3.2cm, height=4.2cm,
      ybar, 
      enlarge x limits=0.5,
      bar width=8pt,
      grid style=dashed,
      ymajorgrids=true,
      xmajorgrids=true,
      ylabel near ticks,
      xlabel near ticks,
      symbolic x coords={De$\rightarrow$En},
      xtick=data,  
      nodes near coords,
      ymin=34.5,ymax=35.5,
      every node near coord/.append style={
        font=\scriptsize,
        rotate=90, 
        anchor=west, 
      }
      ]  
    \addplot[draw=lyyred,fill=lyyred!50] coordinates {(De$\rightarrow$En, 34.65)};
    \addplot[draw=lyyblue,fill=lyyblue!50] coordinates {(De$\rightarrow$En, 35.12)};
    \addplot[draw=lyygreen,fill=lyygreen!50] coordinates {(De$\rightarrow$En, 35.13)};
    \addplot[draw=orange,fill=orange!50] coordinates {(De$\rightarrow$En, 34.89)};
    \addplot[draw=purple!60,fill=lypurple] coordinates {(De$\rightarrow$En, 34.88)};
    \end{axis}  
  \end{tikzpicture}  
  \begin{tikzpicture}  
    \begin{axis}  
    [  
      width=3.2cm, height=4.2cm,
      ybar, 
      enlarge x limits=0.5,
      bar width=8pt,
      grid style=dashed,
      ymajorgrids=true,
      xmajorgrids=true,
      ylabel near ticks,
      xlabel near ticks,
      symbolic x coords={Ru$\rightarrow$En},
      xtick=data,  
      nodes near coords,
      ymin=34.0, ymax=36.5,
      every node near coord/.append style={
        font=\scriptsize,
        rotate=90, 
        anchor=west, 
      }
      ]  
    \addplot[draw=lyyred,fill=lyyred!50] coordinates {(Ru$\rightarrow$En, 34.65)};
    \addplot[draw=lyyblue,fill=lyyblue!50] coordinates {(Ru$\rightarrow$En, 35.70)};
    \addplot[draw=lyygreen,fill=lyygreen!50] coordinates {(Ru$\rightarrow$En, 35.45)};
    \addplot[draw=orange,fill=orange!50] coordinates {(Ru$\rightarrow$En, 35.57)};
    \addplot[draw=purple!60,fill=lypurple] coordinates {(Ru$\rightarrow$En, 35.31)};
    \end{axis}  
  \end{tikzpicture}  
  \begin{tikzpicture}  
    \begin{axis}  
    [  
      width=3.2cm, height=4.2cm,
      ybar, 
      enlarge x limits=0.5,
      bar width=8pt,
      grid style=dashed,
      ymajorgrids=true,
      xmajorgrids=true,
      ylabel near ticks,
      xlabel near ticks,
      symbolic x coords={Cs$\rightarrow$En},
      xtick=data,  
      nodes near coords,
      ymin=29.9, ymax=31.5,
      every node near coord/.append style={
        font=\scriptsize,
        rotate=90, 
        anchor=west, 
      }
      ]  
    \addplot[draw=lyyred,fill=lyyred!50] coordinates {(Cs$\rightarrow$En, 30.00)};
    \addplot[draw=lyyblue,fill=lyyblue!50] coordinates {(Cs$\rightarrow$En, 30.80)};
    \addplot[draw=lyygreen,fill=lyygreen!50] coordinates {(Cs$\rightarrow$En, 30.82)};
    \addplot[draw=orange,fill=orange!50] coordinates {(Cs$\rightarrow$En, 30.69)};
    \addplot[draw=purple!60,fill=lypurple] coordinates {(Cs$\rightarrow$En, 30.26)};
    \end{axis}  
  \end{tikzpicture}  
  \begin{tikzpicture}  
    \begin{axis}  
    [  
      width=3.2cm, height=4.2cm,
      ybar, 
      enlarge x limits=0.5,
      bar width=8pt,
      grid style=dashed,
      ymajorgrids=true,
      xmajorgrids=true,
      ylabel near ticks,
      xlabel near ticks,
      ylabel={BLEU [\%]},
      symbolic x coords={En$\rightarrow$De},
      xtick=data,  
      nodes near coords,
      ymin=31,
      ymax=33,
      every node near coord/.append style={
        font=\scriptsize,
        rotate=90, 
        anchor=west, 
      }
      ]  
    \addplot[draw=lyyred,fill=lyyred!50] coordinates {(En$\rightarrow$De, 31.59)};
    \addplot[draw=lyyblue,fill=lyyblue!50] coordinates {(En$\rightarrow$De, 32.39)};
    \addplot[draw=lyygreen,fill=lyygreen!50] coordinates {(En$\rightarrow$De, 32.16)};
    \addplot[draw=orange,fill=orange!50] coordinates {(En$\rightarrow$De, 32.27)};
    \addplot[draw=purple!60,fill=lypurple] coordinates {(En$\rightarrow$De, 32.13)};
    \end{axis}  
  \end{tikzpicture}  
  \begin{tikzpicture}  
    \begin{axis}  
    [  
      width=3.2cm, height=4.2cm,
      ybar, 
      enlarge x limits=0.5,
      bar width=8pt,
      grid style=dashed,
      ymajorgrids=true,
      xmajorgrids=true,
      ylabel near ticks,
      xlabel near ticks,
      symbolic x coords={En$\rightarrow$Ru},
      xtick=data,  
      nodes near coords,
      ymin=31.0, ymax=33.5,
      every node near coord/.append style={
        font=\scriptsize,
        rotate=90, 
        anchor=west, 
      }
      ]  
    \addplot[draw=lyyred,fill=lyyred!50] coordinates {(En$\rightarrow$Ru, 31.38)};
    \addplot[draw=lyyblue,fill=lyyblue!50] coordinates {(En$\rightarrow$Ru, 32.50)};
    \addplot[draw=lyygreen,fill=lyygreen!50] coordinates {(En$\rightarrow$Ru, 32.52)};
    \addplot[draw=orange,fill=orange!50] coordinates {(En$\rightarrow$Ru, 32.54)};
    \addplot[draw=purple!60,fill=lypurple] coordinates {(En$\rightarrow$Ru, 32.28)};
    \end{axis}  
  \end{tikzpicture}  
  \begin{tikzpicture}  
    \begin{axis}  
    [  
      width=3.2cm, height=4.2cm,
      ybar, 
      enlarge x limits=0.5,
      bar width=8pt,
      grid style=dashed,
      ymajorgrids=true,
      xmajorgrids=true,
      ylabel near ticks,
      xlabel near ticks,
      symbolic x coords={En$\rightarrow$Cs},
      xtick=data,  
      nodes near coords,
      ymin=28.0, ymax=29.2,
      every node near coord/.append style={
        font=\scriptsize,
        rotate=90, 
        anchor=west, 
      }
      ]  
    \addplot[draw=lyyred,fill=lyyred!50] coordinates {(En$\rightarrow$Cs, 28.18)};
    \addplot[draw=lyyblue,fill=lyyblue!50] coordinates {(En$\rightarrow$Cs, 28.74)};
    \addplot[draw=lyygreen,fill=lyygreen!50] coordinates {(En$\rightarrow$Cs, 28.64)};
    \addplot[draw=orange,fill=orange!50] coordinates {(En$\rightarrow$Cs, 28.79)};
    \addplot[draw=purple!60,fill=lypurple] coordinates {(En$\rightarrow$Cs, 28.64)};
    \end{axis}  
  \end{tikzpicture}  
  \begin{tikzpicture}  
    \begin{axis}  
    [  
      width=3.2cm, height=4.2cm,
      ybar, 
      enlarge x limits=0.5,
      bar width=8pt,
      grid style=dashed,
      ymajorgrids=true,
      xmajorgrids=true,
      ylabel near ticks,
      xlabel near ticks,
      symbolic x coords={De$\rightarrow$En},
      xtick=data,  
      nodes near coords,
      ymin=35.5,ymax=35.9,
      every node near coord/.append style={
        font=\scriptsize,
        rotate=90, 
        anchor=west, 
      }
      ]  
    \addplot[draw=lyyred,fill=lyyred!50] coordinates {(De$\rightarrow$En, 35.52)};
    \addplot[draw=lyyblue,fill=lyyblue!50] coordinates {(De$\rightarrow$En, 35.72)};
    \addplot[draw=lyygreen,fill=lyygreen!50] coordinates {(De$\rightarrow$En, 35.76)};
    \addplot[draw=orange,fill=orange!50] coordinates {(De$\rightarrow$En, 35.71)};
    \addplot[draw=purple!60,fill=lypurple] coordinates {(De$\rightarrow$En, 35.76)};
    \end{axis}  
  \end{tikzpicture}   
  \begin{tikzpicture}  
    \begin{axis}  
    [  
      width=3.2cm, height=4.2cm,
      ybar, 
      enlarge x limits=0.5,
      bar width=8pt,
      grid style=dashed,
      ymajorgrids=true,
      xmajorgrids=true,
      ylabel near ticks,
      xlabel near ticks,
      symbolic x coords={Ru$\rightarrow$En},
      xtick=data,  
      nodes near coords,
      ymin=32.0, ymax=34.5,
      every node near coord/.append style={
        font=\scriptsize,
        rotate=90, 
        anchor=west, 
      }
      ]  
    \addplot[draw=lyyred,fill=lyyred!50] coordinates {(Ru$\rightarrow$En, 32.55)};
    \addplot[draw=lyyblue,fill=lyyblue!50] coordinates {(Ru$\rightarrow$En, 33.55)};
    \addplot[draw=lyygreen,fill=lyygreen!50] coordinates {(Ru$\rightarrow$En, 33.32)};
    \addplot[draw=orange,fill=orange!50] coordinates {(Ru$\rightarrow$En, 33.31)};
    \addplot[draw=purple!60,fill=lypurple] coordinates {(Ru$\rightarrow$En, 33.40)};
    \end{axis}  
  \end{tikzpicture}  
  \begin{tikzpicture}  
    \begin{axis}  
    [  
      width=3.2cm, height=4.2cm,
      ybar, 
      enlarge x limits=0.5,
      bar width=8pt,
      grid style=dashed,
      ymajorgrids=true,
      xmajorgrids=true,
      ylabel near ticks,
      xlabel near ticks,
      symbolic x coords={Cs$\rightarrow$En},
      xtick=data,  
      nodes near coords,
      ymin=33.0, ymax=34.2,
      every node near coord/.append style={
        font=\scriptsize,
        rotate=90, 
        anchor=west, 
      }
      ]  
    \addplot[draw=lyyred,fill=lyyred!50] coordinates {(Cs$\rightarrow$En, 33.21)};
    \addplot[draw=lyyblue,fill=lyyblue!50] coordinates {(Cs$\rightarrow$En, 33.50)};
    \addplot[draw=lyygreen,fill=lyygreen!50] coordinates {(Cs$\rightarrow$En, 33.77)};
    \addplot[draw=orange,fill=orange!50] coordinates {(Cs$\rightarrow$En, 33.80)};
    \addplot[draw=purple!60,fill=lypurple] coordinates {(Cs$\rightarrow$En, 33.50)};
    \end{axis}  
  \end{tikzpicture}  
  \caption{Results on WMT17 translation tasks (The first/second row shows results on the test/validation sets.).}
  \label{fig:main_result_wmt17}
\end{figure*}

\subsection{Hyperparameter Tuning vs. Convergence}
To achieve a similar convergence state, in this section, we tune the training hyperparameters of learning rate, warmup steps, training batches, and training steps. 
$L_2$ regularization is also applied for training stability. 

\textbf{Tuning hyperparameters helps the original Transformer model perform better.}
Inspired by \citet{DBLP:conf/wmt/OttEGA18}, here we adopt the larger learning rate ($0.002$ vs. $0.001$), the larger warmup steps (16,000 vs. 8,000), and the larger batch size (8,192 tokens vs. 4,096 tokens) for all of the Transformer-base/big/deep "tuning hyperparameters" models.
For Transformer-big models, we additionally adopt the $L2$ regularization for stable training.
For training steps, the model with hyperparameter tuning has the same or fewer steps than the original model (100K vs. 100K on Transformer-base/deep, 240K vs. 300K on Transformer-big).
As shown in \fig{fig:curve2}, with an optimized learning rate schedule, the original model can reach a convergence state close to that of the parameter sharing model even if the latter uses the same training hyperparameters.
Results in \tab{tab:tune_SIL} further verify the phenomenon in \fig{fig:curve2}.
Although the SIL with hyperparameter tuning (+$L_2$ regularization for training stability) can outperform the original SIL by 0.42 BLEU points, it performs about the same as the original model with hyperparameter tuning.

\textbf{$L_2$ regularization helps to converge stably.}
As shown in \tab{tab:main_result_big}, due to larger learning rates or sharing parameters for multiple times, the loss exploding problem will occur when parameter sharing is applied to Transformer-big models and models with hyperparameter tuning. 
In order to solve the above problems, this paper adopts $L_2$ regularization applied to weight (also called weight decay).
It adds the squared magnitude of the network weights as the penalty term to the original loss function:
\begin{equation}
    \mathcal{L}_{\lambda}(w)=\mathcal{L}(w)+\lambda\norm{w}_2^2
\end{equation}
where $\mathcal{L}$ is the original cross-entropy loss, $\lambda$ is the penalty factor that we set to $0.02$.
This penalty term will encourage the network weights $w$ to be smaller.
Experiments in \tab{tab:main_result_big} show that with the $L_2$ penalty, all experiments can converge stably.


\begin{table}[t!]
  \centering
  \setlength{\tabcolsep}{0.15cm}
  \small
  \renewcommand\arraystretch{1.1}
  \begin{tabular}{l|r|r|r|r|r|r}
      \hline 
      \multicolumn{1}{c|}{\multirow{2}*{Lang.}} & 
      \multicolumn{2}{c|}{Train} & 
      \multicolumn{2}{c|}{Valid} & 
      \multicolumn{2}{c}{Test} \\
      \cline{2-7}
      &
      \multicolumn{1}{c|}{sent.} & 
      \multicolumn{1}{c|}{word} &
      \multicolumn{1}{c|}{sent.} & 
      \multicolumn{1}{c|}{word} &
      \multicolumn{1}{c|}{sent.} &
      \multicolumn{1}{c}{word} \\
      \hline
      \href{https://statmt.org/wmt17/translation-task.html}{En$\leftrightarrow$De} & 5.9M & 276M & 8171 & 356K & 3004 & 128K \\
      \hline
      \href{https://statmt.org/wmt17/translation-task.html}{En$\leftrightarrow$Ru} & 25M & 1.2B & 8819 & 391K & 3001 & 132K \\
      \hline
      \href{https://statmt.org/wmt17/translation-task.html}{En$\leftrightarrow$Cs} & 52M & 1.2B & 8658 & 354K & 3005 & 118K \\
      \hline
  \end{tabular}
  \caption{Data statistics (\# of sentences and \# of words).}
  \label{tab:data}
\end{table}

\subsection{Experiments in Section \ref{sec:convergence}}
\label{sub-sec:experiments-section4}

For more convincing experiments, besides WMT14 En-De and WMT14 En-Fr, we additionally evaluate our method on 6 WMT17 tasks in \fig{fig:main_result_wmt17}.
We use \emph{newstest2017} as the test set and use the concatenation of all available preprocessed validation sets as the validation set for all WMT17 datasets. 
Detailed data statistics as well as the URLs are shown in \tab{tab:data} and Appendix \ref{sec:appendixa}. 

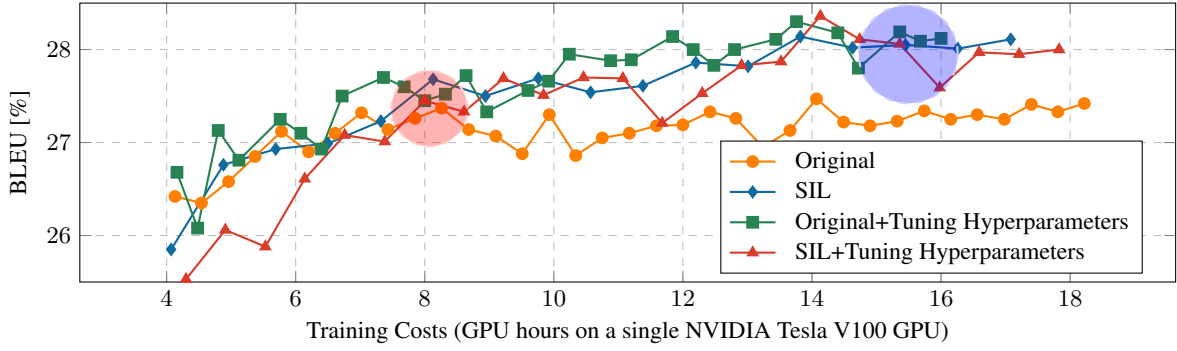
\begin{figure*}[t!]
  \centering
  \small
  \begin{tikzpicture}
      \begin{axis}[
        width=1.0\linewidth,height=0.33\linewidth,
        yticklabel style={/pgf/number format/fixed,/pgf/number format/precision=1},
        ymin=25.5,ymax=28.5,
        ytick={26.0,...,28.0},
        ylabel={BLEU [\%]},
        ylabel near ticks,
        xlabel={Training Costs (GPU hours on a single NVIDIA Tesla V100 GPU)},
        xlabel near ticks,
        xmajorgrids=true,
        ymajorgrids=true,
        grid style=dashed,
        every tick label/.append style={font=\small},
        label style={font=\small},
        legend style={
              column sep=5pt,
              legend columns=1,
        },
        legend cell align={left},
        legend pos={south east},
        ]
        \addplot [orange,thick,mark=*] coordinates {
          (4.13,26.42) (4.54,26.35) (4.96,26.58) (5.37,26.85) (5.78,27.12) (6.20,26.90) (6.61,27.10) (7.02,27.32) (7.43,27.14) (7.85,27.26) (8.26,27.37) (8.68,27.14) (9.10,27.07) (9.51,26.88) (9.93,27.30) (10.34,26.86) (10.75,27.05) (11.17,27.10) (11.59,27.18) (12.00,27.19) (12.42,27.33) (12.82,27.26) (13.24,26.95) (13.66,27.13) (14.07,27.47) (14.49,27.22) (14.90,27.18) (15.32,27.23) (15.74,27.34) (16.15,27.25) (16.56,27.30) (16.98,27.25) (17.40,27.41) (17.81,27.33) (18.22,27.42) 
        };\addlegendentry{Original}
        \addplot [lyyblue,thick,mark=diamond*] coordinates {
          (4.07,25.85) (4.88,26.76) (5.69,26.93) (6.50,26.99) (7.32,27.23) (8.13,27.68) (8.94,27.50) (9.76,27.69) (10.57,27.54) (11.38,27.61) (12.20,27.86) (13.01,27.82) (13.82,28.14) (14.63,28.02) (15.45,28.05) (16.26,28.01) (17.08,28.11)
        };\addlegendentry{SIL}
        \addplot [lyygreen,thick,mark=square*] coordinates {
          (4.16,26.68) (4.48,26.08) (4.80,27.13) (5.12,26.81) (5.76,27.25) (6.08,27.10) (6.40,26.93) (6.72,27.50) (7.36,27.70) (7.68,27.60) (8.00,27.45) (8.32,27.52) (8.64,27.72) (8.96,27.33) (9.60,27.56) (9.92,27.66) (10.24,27.95) (10.88,27.88) (11.20,27.89) (11.84,28.14) (12.16,28.00) (12.48,27.83) (12.80,28.00) (13.44,28.11) (13.76,28.30) (14.40,28.18) (14.72,27.80) (15.36,28.19) (15.68,28.09) (16.00,28.12)
        };\addlegendentry{Original+Tuning Hyperparameters}
        \addplot [lyyred,thick,mark=triangle*] coordinates {
          (4.30,25.53) (4.91,26.06) (5.53,25.88) (6.14,26.61) (6.76,27.08) (7.38,27.01) (8.00,27.45) (8.61,27.33) (9.22,27.69) (9.84,27.51) (10.46,27.70) (11.07,27.69) (11.68,27.21) (12.30,27.53) (12.91,27.83) (13.52,27.87) (14.13,28.36) (14.74,28.11) (15.36,28.06) (15.98,27.59) (16.59,27.97) (17.21,27.95) (17.83,28.00) 
        };\addlegendentry{SIL+Tuning Hyperparameters}
      \end{axis}
      \draw[color=red!20,fill=red,fill opacity=0.3,thick,dashed] (4.6,2.3) circle (0.5);
      \draw[color=blue!20,fill=blue,fill opacity=0.3,thick,dashed] (10.9,3.02) circle (0.65);
  \end{tikzpicture}
  \caption{Performance (BLEU) vs. training costs (measured by GPU hours) on WMT14 En-De.}
  \label{fig:cost}
\end{figure*}

\textbf{The success of parameter sharing depends largely on the optimized training convergence.}
All Transformer-base/deep/big experiments in \tab{tab:main_result_ende}, \tab{tab:main_result_big} and \fig{fig:main_result_wmt17} show that, compared to SIL, SIB, and SIM, models with hyperparameters tuning can also achieve competitive performance.
On some tasks, there is still a small gap between the results of hyperparameter tuning and parameter sharing ($\Delta\mathrm{BLEU}<0.5$), such as Cs$\rightarrow$En in \fig{fig:main_result_wmt17}.
This performance gap can be attributed to model complexity as discussed in Section \ref{sec:complexity}.
Combining Section \ref{sec:complexity} and Section \ref{sec:convergence}, we show that the success of parameter sharing can be largely attributed to better convergence during training rather than the increased model complexity.


\section{Analysis}

To better understand the above conclusions, we conduct a series of analysis mainly on the WMT14 En-De dataset.
For the clarity of these tables and figures, we mainly choose SIL for parameter sharing experiments.

\begin{table}[t!]
  \centering
  \setlength{\tabcolsep}{0.05cm}
  \small
  \renewcommand\arraystretch{1.0}
  \begin{tabular}{l|c|c|c|c|c|c|c|c}
      \hline 
      \multicolumn{1}{c|}{\multirow{2}*{System}} & 
      \multicolumn{2}{c|}{Enc} & 
      \multicolumn{2}{c|}{Dec} & 
      \multicolumn{2}{c|}{BLEU} & \multicolumn{1}{c|}{\multirow{2}*{FLOPs}} & 
      \multicolumn{1}{c}{\multirow{2}*{Costs}} \\
      \cline{2-7}
      &
      \multicolumn{1}{c|}{D} & 
      \multicolumn{1}{c|}{S} &
      \multicolumn{1}{c|}{D} & 
      \multicolumn{1}{c|}{S} &
      \multicolumn{1}{c|}{Test} &
      \multicolumn{1}{c|}{Valid} & \\
      \hline
      Transformer-base & 6 & - & 6 & - & 27.56 & 25.94 & 1.81G & 8.68 \\
      \hline
      Original-1l (25M) & 1 & - & 1 & - & 21.42 & 20.87 & 0.71G & 2.80 \\
      + SIL-share2 & 1 & 2 & 1 & 2 & 22.01 & 21.34 & 0.93G & 3.92 \\
      + SIL-share4 & 1 & 4 & 1 & 4 & 19.88 & 18.70 & 1.37G & 6.44 \\
      + SIL-share6 & 1 & 6 & 1 & 6 & 16.20 & 16.78 & 1.81G & 8.68 \\
      + Tuning Hyper. & 1 & - & 1 & - & \colorbox{gray!10}{\underline{\textbf{22.27}}} & \colorbox{gray!10}{\underline{\textbf{21.43}}} & 0.71G & 5.33 \\
      \hline
      Original-2l (33M) & 2 & - & 2 & - & 25.05 & 24.14 & 0.93G & 3.92 \\
      + SIL-share2 & 2 & 2 & 2 & 2 & 25.13 & 23.70 & 1.37G & 6.44 \\
      + SIL-share4 & 2 & 4 & 2 & 4 & 22.69 & 21.55 & 2.25G & 10.92 \\
      + SIL-share6 & 2 & 6 & 2 & 6 & 19.58 & 18.68 & 3.13G & 21.56 \\
      + Tuning Hyper. & 1 & - & 1 & - & \colorbox{gray!10}{\underline{\textbf{25.59}}} & \colorbox{gray!10}{\underline{\textbf{24.63}}} & 0.93G & 7.67 \\
      \hline
      Original-3l (40M) & 3 & - & 3 & - & 25.95 & 25.02 & 1.15G & 5.04 \\
      + SIL-share2 & 3 & 2 & 3 & 2 & 26.03 & 24.30 & 1.81G & 8.68 \\
      + SIL-share4 & 3 & 4 & 3 & 4 & 23.35 & 22.32 & 3.13G & 21.56 \\
      + SIL-share6 & 3 & 6 & 3 & 6 & 19.18 & 17.99 & 4.46G & 22.68 \\
      + Tuning Hyper. & 3 & - & 3 & - & \colorbox{gray!10}{\underline{\textbf{26.74}}} & \colorbox{gray!10}{\underline{\textbf{25.43}}} & 1.15G & 9.67 \\
      \hline
  \end{tabular}
  \caption{Experiments of various depths (D) of both encoder (Enc) and decoder (Dec) on WMT14 En-De. For SIL we also test the effect of different numbers for parameter sharing (S).}
  \label{tab:shape_encoder_decoder}
\end{table}

\subsection{Experiments on Smaller Models}

Previous experiments have focused on parameter sharing on the encoder, which is due to the fact that the model performance is more sensitive to the encoder.
Here, we shrink the model depth on both the encoder and decoder to compare the approaches of parameter sharing and hyperparameter tuning on smaller models.
In particular, we build three smaller baselines named original-1l, original-2l, and original-3l of different sizes.
Each baseline uses a layer 1/2/3 encoder and a layer 1/2/3 decoder, respectively.
Based on the three smaller baselines, we share the encoder and decoder 2/4/6 times in SIL (named SIL-share2/4/6).

As shown in \tab{tab:shape_encoder_decoder}, the model with hyperparameter tuning consistently outperforms the best SIL model (SIL-share2) by 0.26$\sim$0.71 BLEU points at various model sizes and can save about 1.57$\times$ FLOPs.
To our surprise, SIL does not perform well on smaller models, with only SIL-share2 showing a slight improvement in performance, while SIL-share4 and SIL-share6 drop significantly.
It suggests that choosing an appropriate number to share parameters is important, especially when parameters are very limited.

\subsection{Training Efficiency Analysis}

To compare the average training costs for each approach, \fig{fig:cost} shows the actual runtime and the corresponding BLEU scores (The original model converges near the red circle {\color{red!50}{$\CIRCLE$}}, other models converge near the blue circle {\color{blue!50}{$\CIRCLE$}}.). 
One can observe that although the performance of the original model {\color{orange}{$\CIRCLE$}} cannot be improved as the training time increases, the original model with hyperparameter tuning {\color{lyygreen}{$\blacksquare$}} can achieve almost the same performance as SIL {\color{lyyblue}{$\blacklozenge$}}.
To achieve convergence, Original+Tuning Hyperparameters {\color{lyygreen}{$\blacksquare$}}, SIL {\color{lyyblue}{$\blacklozenge$}}, and SIL+Tuning Hyperparameters {\color{lyyred}{$\blacktriangle$}} take approximately the same time, but the model complexity of the original model {\color{orange}{$\CIRCLE$}}{\color{lyygreen}{$\blacksquare$}} is only half that of the parameter sharing model {\color{lyyblue}{$\blacklozenge$}}{\color{lyyred}{$\blacktriangle$}} (1.81G vs. 3.51G as shown in\tab{tab:main_result_ende}).
In other words, the parameter sharing model and our model can achieve similar performance with the same training costs, but the model complexity of the latter is only half of the former.

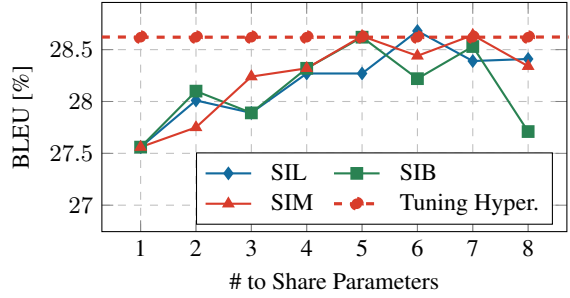
\begin{figure}[t!]
  \centering
  \small
  \begin{tikzpicture}
      \begin{axis}[
        width=1.0\linewidth,height=0.6\linewidth,
        yticklabel style={/pgf/number format/fixed,/pgf/number format/precision=1},
        ymin=27,ymax=28.7,
        enlarge y limits=0.15,
        ylabel={BLEU [\%]},
        ylabel near ticks,
        xlabel={\# to Share Parameters},
        xlabel near ticks,
        symbolic x coords={1,2,3,4,5,6,7,8},
        xmajorgrids=true,
        ymajorgrids=true,
        grid style=dashed,
        xtick=data,
        every tick label/.append style={font=\small},
        label style={font=\small},
        legend style={
              column sep=5pt,
              legend columns=2,
        },
        legend cell align={left},
        legend pos=south east,
        ]
        \addplot [lyyblue,thick,mark=diamond*] coordinates {
            (1,27.56) (2,28.01) (3,27.89) (4,28.27) (5,28.27) (6,28.68) (7,28.39) (8,28.41)
        };\addlegendentry{SIL}
        \addplot [lyygreen,thick,mark=square*] coordinates {
            (1,27.56) (2,28.10) (3,27.89) (4,28.32) (5,28.62) (6,28.22) (7,28.53) (8,27.71) 
        };\addlegendentry{SIB}
        \addplot [lyyred,thick,mark=triangle*] coordinates {
            (1,27.56) (2,27.75) (3,28.24) (4,28.32) (5,28.63) (6,28.44) (7,28.64) (8,28.34)
        };\addlegendentry{SIM}
        \addplot
        [lyyred,very thick,dashed,mark=*] coordinates {
            (1,28.62) (2,28.62) (3,28.62) (4,28.62) (5,28.62) (6,28.62) (7,28.62) (8,28.62)
        };\addlegendentry{Tuning Hyper.}
      \end{axis}
      \draw [lyyred,very thick,dashed] (0.0,2.58) -- (0.5,2.58);
      \draw [lyyred,very thick,dashed] (5.5,2.58) -- (6.14,2.58);
  \end{tikzpicture}
  \caption{Performance (BLEU) vs. the number to share parameters on WMT14 En-De.}
  \label{fig:deeper}
\end{figure}

\subsection{Experiments on Sharing More Times}
\label{sub-sec:comparison}

In \fig{fig:deeper}, we further compare the best performance that can be achieved by all methods.
It can be seen that as the number to share parameters increases, the performance of SIL, SIB, and SIW systems first increases rapidly and then tends to stabilize or even decline.
The highest point of these curves is very close to the model with hyperparameter tuning (the red dotted line above), which further validates our conclusion that the success of parameter sharing depends largely on the optimized training convergence.

\begin{figure}
  \centering
  \small
  \begin{tikzpicture}
    \begin{axis}[
      width=8.0cm, height=3.6cm,
      yticklabel style={/pgf/number format/fixed,/pgf/number format/fixed zerofill,/pgf/number format/precision=0},
      ylabel={Frequency},
      ylabel near ticks,
      xlabel={BLEU [\%]},
      xlabel near ticks,
      enlargelimits=0.1,
      ymax=700,
      symbolic x coords={<10,<20,<30,<40,<50,50+},
      xmajorgrids=true,
      ymajorgrids=true,
      grid style=dashed,
      xtick=data,
      every tick label/.append style={font=\small},
      label style={font=\small},
      ylabel style={yshift=5pt},
      legend style={legend columns=2},
      legend pos=north east,
      legend cell align={left},
    ]
    \addplot [lyyred,thick,mark=square*] coordinates {
      (<10,690) (<20,355) (<30,579) (<40,552) (<50,385) (50+,442) 
    };\addlegendentry{Original}
    \addplot [lyyblue,thick,thick,mark=triangle*] coordinates {
      (<10,671) (<20,333) (<30,569) (<40,542) (<50,384) (50+,504) 
    };\addlegendentry{Tuning Hyper.}           
    \end{axis}
  \end{tikzpicture}
  \caption{The frequency ($\times 10^5$) of BLEU score [\%].}
  \label{fig:error1}
\end{figure}
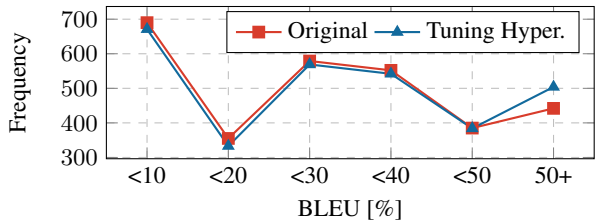

\subsection{Translation Analysis}

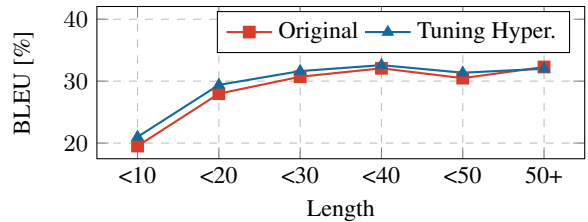
\begin{figure}
  \centering
  \small
  \begin{tikzpicture}
    \begin{axis}[
      width=8.0cm, height=3.6cm,
      yticklabel style={/pgf/number format/fixed,/pgf/number format/fixed zerofill,/pgf/number format/precision=0},
      ylabel={BLEU [\%]},
      ylabel near ticks,
      xlabel={Length},
      xlabel near ticks,
      enlargelimits=0.1,
      ymax=40,
      symbolic x coords={<10,<20,<30,<40,<50,50+},
      xmajorgrids=true,
      ymajorgrids=true,
      grid style=dashed,
      xtick=data,
      every tick label/.append style={font=\small},
      label style={font=\small},
      ylabel style={yshift=5pt},
      legend style={legend columns=2},
      legend pos=north east,
      legend cell align={left},
    ]
    \addplot [lyyred,thick,mark=square*] coordinates {
      (<10,19.52) (<20,27.97) (<30,30.71) (<40,32.07) (<50,30.48) (50+,32.28) 
    };\addlegendentry{Original}
    \addplot [lyyblue,thick,thick,mark=triangle*] coordinates {
      (<10,21.00) (<20,29.39) (<30,31.62) (<40,32.58) (<50,31.36) (50+,32.00) 
    };\addlegendentry{Tuning Hyper.}              
    \end{axis}
  \end{tikzpicture}
  \caption{BLEU score [\%] vs. translation length.}
  \label{fig:error2}
\end{figure}

The Transformer model with hyperparameter tuning always performs better than the original model in the above experiments. 
Here we conduct experiments to better understand this phenomenon. 
We first evaluate the sentence-level BLEU score \footnote{\url{https://www.nltk.org/_modules/nltk/translate/bleu_score.html}} for each translation (3-grams), then group these translations according to their BLEU scores (in \fig{fig:error1}) or lengths (in \fig{fig:error2}).
\fig{fig:error1} shows that the model with hyperparameter tuning has more high-scoring sentences (BLEU [\%]: 50+), which indicates that it does have a better translation ability.
\fig{fig:error2} shows that compared to the original model, the model with hyperparameter tuning translates better on short sentences (Length: <50).
As for long sentences (Length: 50+), these two models have almost the same performance.
This may be due to the fact that long sentences are usually hard ones, and these sentences are often difficult to solve even with improved models.

\section{Related Work}

\subsection{Transformer Varients}

Increasing model complexity can be a straightforward way to improve model performance.
By stacking more layers, deep models have been shown to be effective in recent studies \cite{DBLP:conf/cvpr/HeZRS16,DBLP:journals/corr/ZagoruykoK17,DBLP:journals/corr/Telgarsky15,DBLP:conf/emnlp/LiuLGCH20,DBLP:journals/corr/abs-2008-07772}.
For the Transformer model, \citet{DBLP:conf/acl/WangLXZLWC19} have studied deep encoders and successfully trained a 30-layer Transformer system.
\citet{DBLP:journals/corr/abs-2008-07772} train a 60-layer transformer that outperforms the 6-layer counterpart by 2.5 BLEU points.
\citet{DBLP:journals/corr/abs-2203-00555} prove that the Transformer model can even be trained to 1000 layers.
From another perspective, increasing the model width can also improve Transformer performance.
In addition to increasing the matrix dimension like Transformer-big \cite{DBLP:conf/nips/VaswaniSPUJGKP17}, the multi-branch structure widens the model by implementing multiple parallel computing threads \cite{DBLP:journals/corr/abs-1711-02132,DBLP:journals/corr/abs-2006-10270}.

\subsection{Efficient Methods}

There are several solutions to improve parameter efficiency or computational efficiency in neural networks.
Among them, knowledge distillation treats predictions of the larger teacher network as learned knowledge that can be transferred to the smaller student network \cite{DBLP:journals/corr/HintonVD15,DBLP:journals/corr/FreitagAS17,DBLP:conf/emnlp/KimR16,DBLP:journals/corr/abs-2009-09152}.
Quantization represents the model with lower-precision numbers such as 8-bit integers \cite{DBLP:conf/nips/HubaraCSEB16,DBLP:conf/iclr/MicikeviciusNAD18,DBLP:conf/naacl/QuinnB18,DBLP:conf/ijcai/LinLLXLZ20}.
Our work follows another line of research called parameter sharing to help the model learn more with limited parameters \cite{DBLP:conf/cvpr/GaoDH19,DBLP:conf/iclr/UllrichMW17,DBLP:journals/corr/abs-2106-10002,DBLP:conf/naacl/HaoWYWZT19,DBLP:journals/corr/abs-2104-06022}.



\section{Conclusion}

In this paper, we study the parameter sharing approach from two perspectives: model complexity and training convergence.
Experiments on 8 WMT machine translation tasks show that the success of this approach depends largely on the optimized training convergence and a small part on the increased model complexity.
Inspired by this, we further tune the training hyperparameters for better convergence. 
With this strategy, the original model without parameter sharing can achieve competitive performance against the sharing model, but with only half the model complexity.


\section*{Limitations}

\noindent \textbf{Training Costs.}
We demonstrate that the success of the parameter sharing approach can be largely attributed to its optimization of model convergence and prove that the original model without sharing can also achieve competitive performance by tuning training hyperparameters.
Compared to parameter sharing models, it can save half the model complexity to make the model easier to deploy on mobile devices. It can also save about 30\% training costs.
But compared to the original model, it still requires additional computational costs (the total training time is about $1.5 \times$ of the original model). 

\noindent \textbf{More Possibilities.}
We analyze two factors that may play an important role in the parameter sharing approach: model complexity and training convergence.
In a sense, a neural network can be seen as a black box, and it is very likely that there are other factors that we have not realized.
Other factors hidden inside the black box may require a more microscopic perspective and related work on neural network visualization.

\bibliography{anthology,custom}
\bibliographystyle{acl_natbib}


\appendix

\section{Details of Datasets}
\label{sec:appendixa}

We evaluate our experiments on 8 WMT14 and WMT17 machine translation tasks.

\textbf{WMT14 tasks}.
We choose WMT14 English-German (En-De) and English-French (En-Fr) as the datasets for our main experiments. 
Detailed data statistics as well as the URLs are shown in \tab{tab:data1}.
For the En-De task, we share the source and target vocabularies. 
We choose \emph{newstest-2013} as the validation set and \emph{newstest-2014} as the test set.
For the En-Fr task, we validate the system on the combination of \emph{newstest-2012} and \emph{newstest-2013}, and test it on \emph{newstest-2014}.

\textbf{WMT17 tasks.}
For more convincing experiments, besides WMT14 En-De and WMT14 En-Fr, we additionally evaluate our method on 6 WMT17 tasks.
Detailed data statistics as well as the URLs are shown in \tab{tab:data}. 
All datasets are the official preprocessed version from WMT17 website. 
We use \emph{newstest2017} as the test set and use the concatenation of all available preprocessed validation sets as the validation set for all WMT17 datasets:
\begin{itemize}
    \setlength\itemsep{1pt}
    \item For WMT17 En$\leftrightarrow$De, we use the concatenation of \emph{newstest2014}, \emph{newstest2015} and \emph{newstest2016} as the validation set.
    \item For WMT17 En$\leftrightarrow$Ru, we use the concatenation of \emph{newstest2014}, \emph{newstest2015} and \emph{newstest2016} as the validation set.
    \item For WMT17 En$\leftrightarrow$Cs, we use the concatenation of \emph{newstest2014}, \emph{newstest2015} and \emph{newstest2016} as the validation set.
\end{itemize}
 
We tokenize every sentence using the script in the Moses toolkit and segment every word into subword units using Byte-Pair Encoding \cite{DBLP:conf/acl/SennrichHB16a}. 
BPE with 32K merge operations is applied to all these datasets.
In addition, we remove sentences with more than 250 subword units \cite{DBLP:conf/acl/XiaoZZL12} and evaluate the results using multi-bleu.perl \footnote{\url{https://github.com/moses-smt/mosesdecoder/blob/master/scripts/generic/multi-bleu.perl}}.

\section{Details of Model Setups}

Our baseline system is based on the open-source implementation of the Transformer model presented in \citet{DBLP:conf/naacl/OttEBFGNGA19}'s work.
We experiment with the Transformer-base and Transformer-deep settings in \tab{tab:main_result_ende} and experiment with the Transformer-big in \tab{tab:main_result_big}.
The standard Transformer-base and Transformer-big systems both consist of a 6-layer encoder and a 6-layer decoder.
The Transformer-deep system has 12 layers.
The embedding size (width) is set to 512 for Transformer-base/deep and 1,024 for Transformer-big. 
The FFN hidden size equals to 4$\times$ embedding size in all settings.
Dropout with the value of 0.1 is used in Transformer-base/deep and 0.3 in Transformer-big.
We also adopt the relative positional representation (RPR) \cite{DBLP:conf/naacl/ShawUV18} to construct stronger baselines.
For all models, we use Adam optimizer with $\beta_1=0.9$ and $\beta_2=0.997$, we stop training until the model stops improving on the validation set. 
All experiments reported in this paper are trained on 8 NVIDIA Tesla V100 GPUs with mixed-precision training \cite{DBLP:conf/iclr/MicikeviciusNAD18} and tested on the model ensemble by averaging the last 5 checkpoints.

\end{document}